# Robust retinal identity matching for patient identity verification and retrieval in large ophthalmic databases

Jose D. Vargas Quiros[1,2], Dennis Bontempi[5,6], Jeroen Vermeulen[1,2], Mattia Tomasoni[8,9], Bart Liefers[1,2], Sven Bergmann[5,6,7] and Caroline C.W. Klaver[1,2,3,4]

[1]**Department of Ophthalmology, Erasmus University Medical Center, Rotterdam, the Netherlands**

[2]**Department of Epidemiology, Erasmus University Medical Center, Rotterdam, the Netherlands**

[3]**Department of Ophthalmology, Radboud University Medical Center, Nijmegen, the Netherlands**

[4]**Institute of Molecular and Clinical Ophthalmology, University of Basel, Switzerland**

[5]**Department of Computational Biology, University of Lausanne, Lausanne, Switzerland**

[6]**Swiss Institute of Bioinformatics, Lausanne, Switzerland**

[7]**Department of Integrative Biomedical Sciences, University of Cape Town, Cape Town, South Africa**

[8]**Department of Ophthalmology, University of Lausanne, Fondation Asile des Aveugles, Jules Gonin Eye Hospital, Lausanne, Switzerland**

[9]**Platform for Research in Ocular Imaging, Fondation Asile des Aveugles, Jules Gonin Eye Hospital, Lausanne, Switzerland.**

**Corresponding author:** Jose D. Vargas Quiros

**Abstract:** Patient identity errors affect the integrity of longitudinal medical imaging records, but retinal photographs contain distinctive anatomical patterns that can be used to match them. We demonstrate, for the first time, that an AI-based retinal identity-matching system trained on a diverse longitudinal dataset is capable of highly accurate identity verification and retrieval across imaging devices, decades of follow-up, and independent cohorts. In the Rotterdam Study, UK Biobank, and Age-Related Eye Disease Study, verification missed only 0.4, 9.3, and 1.1 incorrect assignments per 1,000, respectively, at a fixed rate of five false alerts per 1,000 correct assignments. Retrieval ranked the correct identity first for 99.7%, 97.2%, and 97.6% of queries, respectively, across databases containing thousands of identities. Expert-assisted auditing also uncovered mislabeled images in all three cohorts. Our findings demonstrate the potential of retinal identity matching for reliable, scalable verification and record recovery.

# Introduction

Reliable patient identity is a prerequisite for safe digital medicine. Modern healthcare systems and large population databases increasingly depend on longitudinal electronic health records, imaging archives, screening programs, and AI-enabled decision support. These systems assume that each examination, image, diagnosis, and clinical label is correctly linked to the individual to whom it belongs. In practice, however, patient identity errors do occur[1–3], leading to consequences ranging from administrative inconvenience to the exclusion of valid measurements, fragmentation of clinical records, misattribution of findings to the wrong person, pollution of research databases, and downstream decisions based on incomplete or incorrect information[1,4,5]. Imaging databases are no exception because many clinical, screening, and research workflows rely on human entry of a patient identifier at the point of capture[6,7], making them vulnerable to human error[8]. When such errors are discovered retrospectively, correcting mistakes in image identity can require searching across large picture archiving and communication systems (PACS) repositories[2] making systematic correction difficult and often impractical.

However, many medical imaging modalities capture anatomical structures that can serve as biometric identifiers[9,10] because they contain patterns that are highly unique of an organ or individual. This creates an opportunity to use the image itself to detect identity mismatches and recover the correct identity[11]. In radiology, researchers have explored this idea using a range of image-matching approaches, including normalized cross-correlation, hand-crafted image features, and template- or edge-based methods[3,12–14]. In computer vision and pattern recognition, the related task of matching observations to the correct individual is commonly referred to as person re-identification[15–18], and has been extensively studied using faces or body appearance. More recently, deep metric learning has enabled patient identity to be used directly as a supervision signal, allowing models to learn image representations in which images from the same individual are close together and those from different individuals are separated. Such approaches have shown that chest X-rays can support patient re-identification across repeated examinations and substantial time intervals[19,20]. Similar AI-based approaches have also been investigated using MR imaging and trunk CT, including for applications in data curation[21,22]. Together, these studies suggest that medical images can contain sufficient identity information to support automated patient verification and retrieval.

Retinal imaging is a particularly promising setting for this approach. Retinal vascular patterns, visible in commonly used modalities such as color fundus imaging (CFI) and optical coherence tomography (OCT) are described as highly individual and different even between monozygotic twins[23]. At the same time, retinal imaging is no longer confined to specialist ophthalmology clinics. Its use is expanding through national diabetic-eye screening programs[24,25], teleophthalmology[26], portable color fundus cameras, assistive AI systems in primary care, and emerging 'oculomics' applications[27–29]. Despite this growing volume and diversity of retinal imaging, the use of retinal images as biometric identifiers for patient verification and retrieval has received comparatively little attention. Early work explored matching retinal vascular patterns using fuzzy logic and detection of vascular bifurcations and crossings[30,31]. More recently, deep learning foundation models have been investigated for patient retrieval from ophthalmic images, demonstrating that retinal images contain information that can support re-identification[32]. A response to this article argued that the re-identification ability of the model could be explained by the presence of near-duplicate images in the evaluation set, highlighting the importance of dataset composition[33]. However, these studies neither optimized a system for patient re-identification nor evaluated transfer performance across independent datasets or clinically relevant longitudinal scenarios. Reported identity retrieval performance also remains considerably below that achieved in radiological imaging and is further called into question by presence of near-duplicate images in the evaluation sets.

It therefore remains unclear whether AI-based retinal biometrics can provide a robust and generalizable solution to patient identity errors in clinical and research settings. In this study, we address this question by developing and comprehensively evaluating an AI-based system for identity verification and image retrieval in longitudinal ophthalmic cohorts. We use CFIs from the Rotterdam Study (RS), a large population-based cohort containing over 30 years of repeated ophthalmic imaging, which enables training and evaluation across diverse ages, imaging devices, image quality levels, and retinal fields. We then evaluate the performance of our system in two complementary tasks: identity verification, which asks whether an image belongs to the claimed individual, and identity retrieval, which asks which individual in a database an image belongs to. Evaluation is performed under various clinically relevant scenarios that include the removal of near-duplicate images. We make our system available for clinical and research purposes under an End User License Agreement.

# Methods

We developed an image-based system for patient identification and retrieval system with the goal of correcting patient identity errors in research and clinical databases. The foundation of our system is an AI-based identity encoder, trained to produce latent vectors that separate the different patient-eye identities present in the data. Final evaluation in the clinically relevant tasks of patient identity verification and patient identity retrieval was preceded by an identity label expert audit of the evaluation databases, aimed at ensuring accurate ground truth. The following sub-sections detail our data sources, the development of the identity encoder model, the identity audit process, and the evaluation procedures. Figure 1 shows an overview of our system, showcasing the identity encoder and its two evaluated applications.

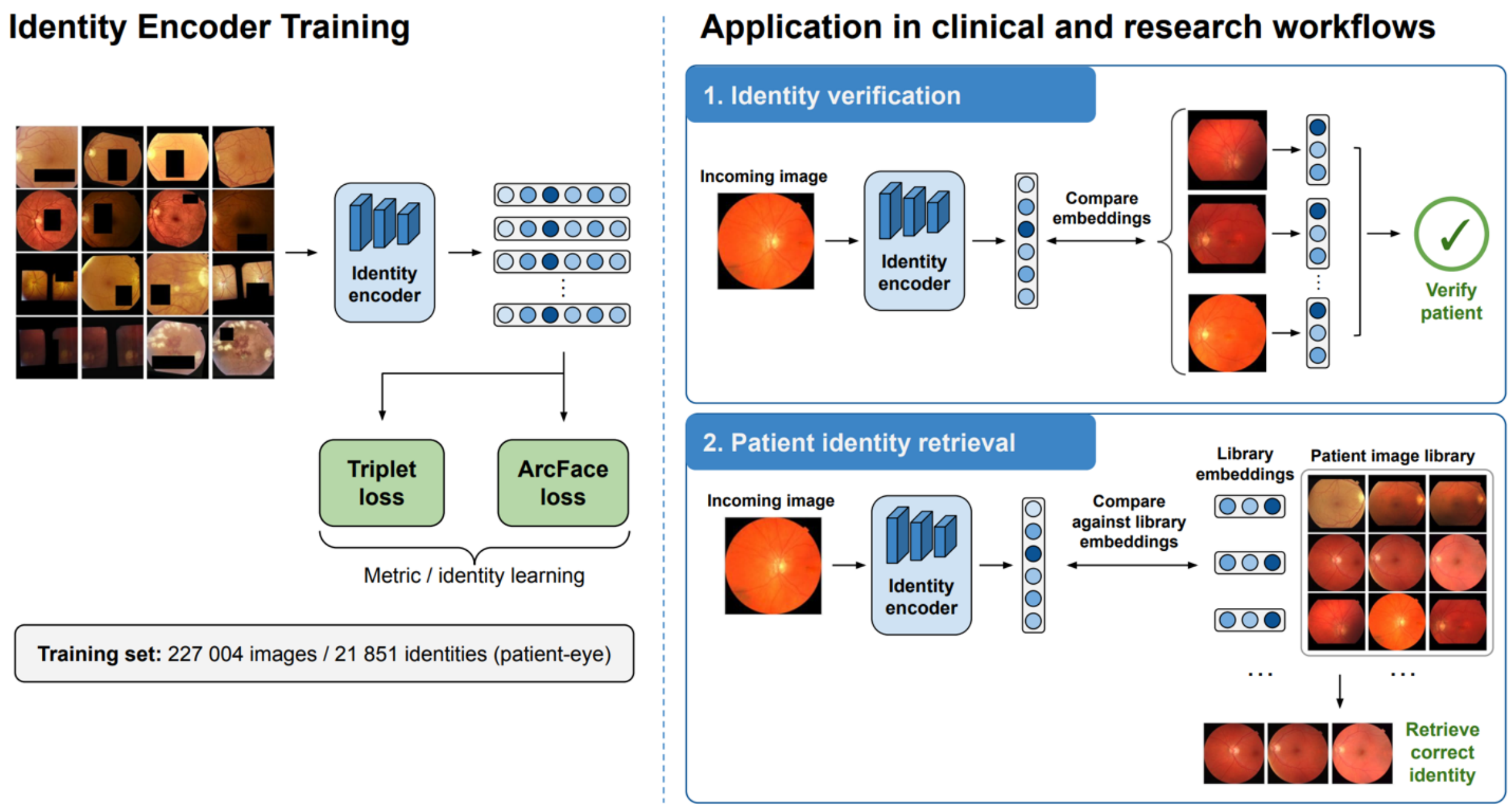

*Figure 1. Overview of our system, which is based on an Identity Encoder with a ConvNextV2 backbone, a combination of Triplet and ArcFace losses, and trained on longitudinal imaging with 21851 patient-eye identities. The encoder was trained to produce identity embeddings, where images from the same patient-eye cluster together in embedding space, while images from distinct patient-eyes are separated. The system is evaluated in two clinical and research applications: 1) Identity verification, where an incoming or target image's embeddings are compared to the embeddings from a patient's existing images to flag wrong identity assignments before they occur, or to retrospectively flag issues in existing databases. 2) Patient identity retrieval, where the goal is to find the correct patient identity given an image.*

## Training and evaluation databases

We made use of large databases with color fundus imaging to train and evaluate our system. The Rotterdam Study (RS) is an ongoing, prospective, population-based cohort study conducted in the Ommoord district of Rotterdam, the Netherlands. It was initiated in 1990 to investigate chronic diseases in middle-aged and older adults. In total, the RS includes approximately 15,000 participants aged ≥45 years. Participants undergo repeated examinations and longitudinal follow-up, including extensive clinical, imaging, genetic, and lifestyle assessments. The RS contains CFIs from various conventional monoscopic fundus cameras as well as stereoscopic CFIs (Topcon Corp., Tokyo, Japan). Both macula and optic disc-centered CFIs were included. The resulting dataset contained 327,519 images from 15,705 participants with a range of follow up between zero (single visit) and 32.61 years. The mean follow-up was 8.6 years, with an average of 2.7 visits per patient, and a maximum of 8.

Because left and right eyes contain distinct anatomical patterns, and attribution mistakes often involve mis-assignment of left-right eye laterality, we defined the identity as the participant-eye combination. In other words, left and right eyes from the same participant are treated as separate identities for training and evaluation purposes. The RS dataset was divided at the identity level into a 70/10/20 split for training, validation and testing / benchmarking. This ensured that no participant-eye identity was shared between the training and held-out test sets. The resulting dataset consisted of 21,851 training identities (227,004 CFIs); 3,123 validation identities (32,477 CFIs); and 6,243 testing / benchmarking identities (65,018 CFIs).

For external evaluation, we used CF imaging from the UK Biobank (UKBB) – a large-scale, population-based prospective cohort of approximately 500,000 participants from the United Kingdom[34] – and from the Age-Related Eye Disease Study (AREDS) – a large, multicenter, longitudinal clinical study designed primarily to investigate age-related macular degeneration and collected in the United States across 11 clinical sites[35].

Of the UKBB cohort, 68,514 participants with mean age 57.3 years (standard deviation of 8.2) underwent retinal imaging between 2009 and 2010 at 6 assessments centers. CFIs in the UK Biobank were acquired using a Topcon 3D OCT-1000 Mark II camera with a 45° field-of-view and are macula-centered[36]. Our evaluation set consisted exclusively of subjects with repeat imaging (two visits), which amounted to 2,254 patients, 4,440 identities, and 8,880 CFIs. In most cases, this set contains a single image per visit per eye. Of these visits, only 33 contain a repeat image, usually due to the first CFI being of poor quality. In these cases, we kept only the latest image to ensure consistency.

The AREDS dataset was collected between 1992 and 1998 from participants aged 55–80 years with varying degrees of AMD who were enrolled and followed longitudinally[35]. CFI images were captured by the Zeiss FF450 camera over a mean follow-up time of 7.2 years, with a maximum of 12 years. The mean number of visits per patient was 7.6, with a maximum of 15. Our evaluation set consists of 4,474 patients, 8,882 identities and 156,853 CFIs. AREDS includes CFIs centered on the macula, the optic disc, and peripheral regions, usually temporal to the macula.

## Identity encoder model development

The foundation of our system was an identity encoder trained as a metric-learning model. Because our goal was to design an AI system for biometric recognition robust to imaging differences, we drew on best practices from computer vision literature on face recognition and person re-identification[17,18,37]. The training objective combined additive angular-margin softmax loss (ArcFace), originally developed for face recognition[37] with triplet loss, which directly encourages an embedding space suited to retrieval-based matching[18]. The triplet loss encouraged embeddings from the same patient-eye identity to be closer than embeddings from different identities by a margin, while the ArcFace loss promoted angular separation between identity classes on the normalized embedding hypersphere.

A ConvNextV2-Tiny architecture initialized from ImageNet-pretrained weights was used as the backbone. This choice was motivated by the strong performance of ConvNeXt-style convolutional backbones across image benchmarks, as well as evidence that ImageNet-pretrained representations transfer well to downstream visual recognition tasks[38,39]. The original classification layer was replaced by an embedding head consisting of a linear projection to a 512-dimensional vector followed by batch normalization. The resulting embedding was L2-normalized, and Euclidean distances in this normalized embedding space were used both during training and at inference for identity verification and retrieval. Images were resized to 384 × 384 pixels and normalized using ImageNet statistics.

We used PK identity sampling together with hard triplet mining, following established practice in person re-identification[18]. PK sampling constructs each training batch by selecting a fixed number of patient identities (P) and multiple images per identity (K), ensuring that each batch contains both same-identity and different-identity examples for metric learning. In our setting, each batch contained 8 identities with 4 images per identity. Hard triplet mining then selected the most challenging examples within the batch—the most dissimilar same-identity image as the positive and the most similar different-identity image as the negative—to focus on learning difficult identity comparisons. Together, these strategies provided informative within-batch comparisons and encouraged representations that are robust to within-patient imaging variation while separating similar-looking patients.

We applied stochastic augmentations designed to preserve identity-relevant retinal structure while improving robustness to acquisition variability. These included random centered circular field-of-view crops, small affine perturbations with translation, scaling, and rotation, photometric augmentation through color jitter, gamma correction, contrast enhancement, and warm yellow-purple color shifts, as well as occasional defocus blur and Gaussian noise. Random rectangular erasing was applied to up to 25% of the image area to reduce reliance on localized artifacts. Supplementary Figure 2 displays a random sample of training and validation identities as presented to the model.

The ConvNeXtV2 backbone was fine-tuned end-to-end using the AdamW optimizer. Models were trained for up to 60 epochs with mixed precision, with hyperparameters and stopping criteria selected based on validation mean average precision (mAP). For inference, the classification head was discarded, and patient identity verification and retrieval were performed using distances between the normalized embeddings.

## Identity label audit of the evaluation databases

Large research datasets such as RS, UKBB, and AREDS likely contain a fraction of identity assignment errors and unusable images. We therefore performed a structured identity-label audit of the evaluation databases before the final performance evaluation. The objectives were to identify images that were assigned to the wrong patient-eye label, identify images with no visible anatomy that would make them impossible to adjudicate, and establish a reliable ground truth for evaluating the model. The audit also provided an opportunity to assess the occurrence of identity mis-adjudication in large ophthalmic imaging databases.

Because a complete manual review of all images would be impractical at the scale of these datasets, we used the identity encoder as a screening tool to identify candidate identity inconsistencies for expert review. Importantly, model-based screening was used only to determine which images warranted further review; the model output was not used as the criterion for determining whether an image was incorrectly labelled. Final decisions were made by ophthalmic researchers experienced in retinal vascular analysis of color fundus images, based on direct visual comparison of retinal anatomy.

For each recorded patient-eye identity containing at least two images, the identity encoder was first used to extract an embedding for every CFI. All pairwise Euclidean distances between images assigned to the same patient-eye identity were then calculated. These distances were used to construct an undirected graph for each identity, in which two images were connected when their embedding distance was below a same-identity verification threshold calibrated on the independent validation set. Connected components were identified within each graph. The largest component was designated as the main cluster and was taken to represent the predominant identity pattern for that patient-eye. Images belonging to smaller disconnected components were flagged as candidate outliers for expert review.

Each candidate outlier was subsequently reviewed alongside images from its corresponding main cluster. The reviewer assessed the retinal anatomy and classified the image into one of four categories: *incorrect identity*, *correct identity*, *anatomy not visible*, and *other (undedidable)*. The review was performed independently of the model output: reviewers were not provided with the model score or distance that had led to the image being flagged.

An image was classified as having an *incorrect identity* when the expert was certain that it captured a different eye from those in the main identity cluster, based primarily on discordant anatomy. Conversely, an image was classified as having the *correct identity* when the expert was certain that it corresponded to the same eye as the main cluster. Images were classified as *anatomy not visible* when the retinal anatomy was not visible in the image, normally due to a failed acquisition. Poor image quality, or blur alone did not qualify for this category when retinal anatomy remained visible. All other cases were classified as *other (undecidable)*, indicating that the identity was undecidable for other reasons. To ensure a fair evaluation of the algorithm, only images classified as incorrectly adjudicated or having no visible anatomy were excluded from the primary evaluation set. Images classified as correctly adjudicated or undecidable for other reasons were retained.

For each dataset, the audit proceeded in rounds. The verification threshold was first calibrated to produce five false alerts per 1000 cases in the independent validation set of the Rotterdam Study, for the same verification task. This tight threshold tended to produce a small number of true positive clusters – real mis-assigned images – in the evaluation set, which were audited manually. After excluding images confirmed by expert review to have incorrect identity assignments or no visible anatomy from the database and excluding all previously reviewed images from being flagged again, the screening and review process was repeated using progressively more permissive thresholds calibrated for 10 and 20 false alerts per 1,000 cases. These thresholds increased the number of candidate images presented for review and thereby allowed additional potential identity errors to be identified. The audit was stopped when further relaxation of the threshold resulted predominantly in false alerts and no longer yielded substantial numbers of confirmed identity corrections.

## Evaluation of patient verification and identity retrieval performance

We evaluated the performance of our identity encoder as two different but related systems. The rationale was to mimic two relevant applications of our system: patient verification, where the (anonymized) identity assignment of an image is verified; and identity retrieval, in which the system attempts to link an image to the correct patient identity.

### Patient verification evaluation

In patient verification, a retinal image has a claimed identity assignment that must be verified. Given a query image and its claimed identity, the system compares the query embedding with the embeddings of other images assigned to the claimed identity and flags the assignment when the images are insufficiently similar. Evaluation required presenting both correct and incorrect assignments to the system. Correct assignments (negative cases) were constructed trivially by using a query image from the same patient-eye identity as the reference set. Incorrect assignments (positive cases) were constructed from the corresponding negative cases by replacing the query image with an image from a different patient, while matching the same laterality (left or right eye), device, and retinal region (macula, optic disc, or other). This sampling strategy was intended to reflect plausible identity mis-assignments while avoiding cases that could be distinguished trivially

We considered two clinically motivated scenarios. In the *Retrospective-only* scenario, only images acquired before the query date were available as references, representing a real-time check against the patient's existing imaging history. In the *Retrospective + Prospective* scenario, all other eligible images from the claimed identity were available, representing verification when the complete longitudinal record can be accessed.

Each scenario was further divided into several sub-scenarios to assess robustness to clinically relevant sources of variation. In the *Same device* scenario, the reference set was restricted to images acquired using the same device as the query, to simulate a database acquired using a single device. In the *Near-duplicates* sub-scenario, near-duplicate images were included in the evaluation, to measure performance on the complete database. Other sub-scenarios were obtained by stratifying or filtering the full database according to retinal region—optic-disc-centered images (F1) and macula-centered images (F2)—or image quality (Q1+Q2 and Q3+Q4 quartiles). Image quality was estimated using a deep learning model trained on the EyePACS EyeQ quality assessment CFI dataset[40,41]. Except for the *Near-duplicates* sub-scenario, all scenarios involved the removal of near-duplicates. Near-duplicates were defined as images of the same retinal region (macula- or optic-disc-centered) acquired during the same visit using the same imaging device. For each such group, we retained only the latest image by timestamp.

For each verification case, the system calculates the Euclidean distance between the query embedding and the embeddings of all eligible reference images. The verification score was defined as the minimum distance to any reference image belonging to the claimed identity. Thus, a small minimum distance indicates consistency with the claimed identity, whereas a large distance indicates a potentially incorrect assignment. Discrimination between correct and incorrect assignments was quantified using the area under the receiver operating characteristic curve (AUROC). To assess performance at clinically relevant operating points, we used thresholds pre-specified and calibrated on the validation set to generate 5, 10 or 15 false alerts per 1,000 correct assignments. At these operating points, we reported the number of false alerts (false positives) and missed incorrect assignments (false negatives).

## Patient identity retrieval evaluation

In our evaluation of patient identity retrieval, the identity of a query image is presumably unknown. The goal of the system is to retrieve the correct identity by comparing the query image's embeddings with those of the full imaging database, known as the gallery. For each query, the system compares its embedding with the embeddings of eligible gallery images and ranks patient identities according to their similarity to the query.

We considered the same two longitudinal scenarios as in the verification evaluation. In the *Retrospective + Prospective* scenario, the gallery consisted of the complete imaging databases from the Rotterdam Study (RS), UK Biobank (UKBB), and AREDS, after removal of near-duplicate images. In the *Retrospective-only* scenario, the gallery was restricted to images acquired before the query image, thereby simulating retrieval using only information that would have been available at the time of the query. Queries for which no eligible prior imaging was available for the relevant evaluation were excluded. The same sub-scenarios were used to assess the effect of imaging devices, retinal regions, and image quality. For the *Same device* scenario, the gallery was restricted to images acquired using the same device as the query. Other scenarios used galleries restricted to macula- or optic-disc-centered images, or to the Q1+Q2 or Q3+Q4 image-quality quartiles.

Retrieval performance was summarized using Recall@1, Recall@5, and mean average precision (mAP). Recall@1 represents the proportion of queries for which the correct identity was ranked first, whereas Recall@5 represents the proportion for which the correct identity appeared among the five highest-ranked identities. Mean average precision summarizes ranking quality across the complete result list by calculating the average precision for each query and then averaging these values across queries. Together, these metrics capture both the accuracy of the highest-ranked identity and the quality of the broader retrieval ranking.

## Analysis of identity embeddings

To characterize the information encoded by the learned identity embeddings, we evaluated their ability to predict imaging-device, demographic, and disease-related variables. The analysis was performed using embeddings generated for CFI images from the Rotterdam Study test set. The variables considered were imaging device, patient sex, age, ethnicity, hypertension, smoking status, age-related macular degeneration (AMD), and glaucoma status. Disease status variables and smoking status were assessed for each RS visit, matching the imaging date. Ophthalmic diseases (AMD and glaucoma) were assessed and linked separately for the left and right eyes.

Each patient-eye identity was represented by a single CFI selected as the image with the highest image-quality score. The resulting dataset was split at the identity level into training, validation, and test sets using a 65%/10%/25% split. For each target variable, an L2-regularized logistic regression classifier was trained using the corresponding identity embedding as input, and a categorical variable as predictive outcome. The regularization strength was selected on the validation set from a predefined grid, and the final classifier was evaluated on the held-out test set. Binary target variables were evaluated using the area under the Area Under the Receiver Operating Characteristic Curve (AUC), while multi-class target variables (ethnicity, imaging device and patient age) were evaluated using multiclass, one-vs-rest AUC.

## Code Availability

Our identity encoder model and related code for patient identity verification and retrieval using the system will be made available to verified parties for clinical and research purposes under an End User License Agreement (EULA). The EULA requires signing parties to provide a clear motivation for use – commercial or non-commercial – and to commit to an adequate handling and responsible use of the model. Please see the Discussion section for related argumentation.

## Use of generative AI tools

During the preparation of this manuscript, the authors used ChatGPT (OpenAI) to assist with manuscript drafting, language refinement, and software development. All AI-generated content and code were reviewed, modified where necessary, and validated by the authors. The authors take full responsibility for the accuracy, integrity, and scientific content of the manuscript.

# Results

We evaluated our system in real, large-scale research databases using longitudinal imaging acquired with multiple devices and under diverse acquisition conditions. Our evaluation extends beyond within-dataset benchmarking by combining expert-reviewed identity auditing, clinically motivated verification and retrieval scenarios, and zero-shot external validation.

## Identity audit of UKBB, AREDS, and Rotterdam Study CFI databases

The manual expert verification of images flagged by the model as potentially assigned to the wrong patient (see Identity label audit of the evaluation databases in the Methods section) was performed on the Rotterdam Study (evaluation set only), AREDS, and UKBB CFI databases. The audit was performed by one expert for the RS and AREDS databases, and a different auditor for the UKBB database. Both experts reported little difficulty assessing flagged cases when the anatomy was visible in both the flagged and reference images. Cases with little overlap between imaged regions – more common in RS and AREDS – often required more time. Figure 2 shows sample flagged and reference images shown to the expert as part of the identity label audit of the RS database, for the four possible review decisions. Equivalent samples are included in the Supplementary Appendix C for AREDS and UKBB.

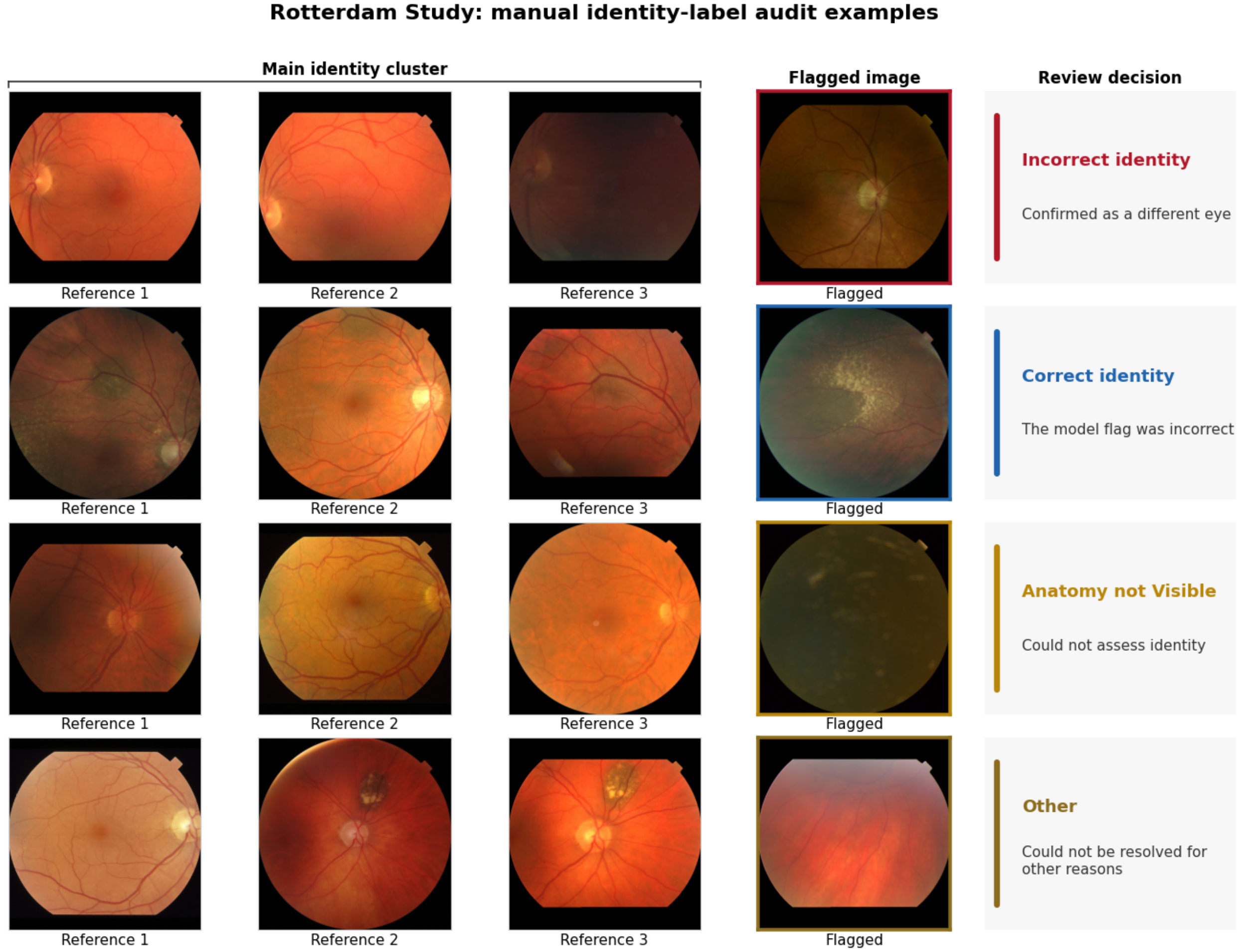

*Figure 2. Samples from the identity label audit of the Rotterdam Study CFIs. The first three columns display images from the majority cluster shown to the human expert. The fourth column (Flagged image) shows the image identified by the system as potentially assigned to the wrong identity. By comparing the vascular patterns in the flagged image with those in the main cluster images, the reviewer decided between confirming an incorrect identity, confirming a correct identity, or an undecidable outcome.*

Table 1 presents the outcomes from the audit of the suggestions produced by the model at increasingly loose thresholds calibrated at 5, 10, and 20 FP/1000 samples, for the three databases. As expected, tighter thresholds tended to produce fewer, more correct candidate images, for which the patient identity was indeed confirmed to be incorrect or undecidable. Initially, a high proportion of flagged images had unrecognizable anatomy due to acquisition failure or extreme poor quality. As the threshold was loosened and the database was cleaned, the number of correct flags and low-quality images decreased until most flags were incorrect or undecidable.

*Table 1. Outcome of the identity-label audit of CFIs from the UKBB, AREDS, and Rotterdam Study population databases. Candidate images indicate the number of images flagged as potentially having a wrong identity at the indicated threshold level.*

*"Incorrect identity" refers to images that were confirmed to have the wrong identity assignment. "Correct identity" refers to images for which the flag was incorrect, as they were correctly assigned. "Anatomy not visible" indicates cases for which no anatomy was visible in the images. "Other (undecidable)" indicates that the expert could not decide for other reasons, such as lack of anatomical overlap.*

| Dataset | Round | Threshold | N images flagged | Incorrect identity | Correct identity | Anatomy not Visible | Other (undecidable) |
|---|---|---|---|---|---|---|---|
| Rotterdam Study | 1 | 5 | 414 | 215 (51.9%) | 1 (0.242%) | 178 (43%) | 20 (4.83%) |
| | 2 | 10 | 424 | 109 (25.7%) | 106 (25%) | 135 (31.8%) | 74 (17.5%) |
| | 3 | 20 | 425 | 39 (9.18%) | 246 (57.9%) | 87 (20.5%) | 53 (12.5%) |
| | Total | | 1,618 | 363 (22.4%) | 708 (43.8%) | 400 (24.7%) | 147 (9.09%) |
| AREDS | 1 | 5 | 934 | 250 (26.8%) | 179 (19.2%) | 241 (25.8%) | 264 (28.3%) |
| | 2 | 10 | 1,765 | 8 (0.453%) | 1,201 (68%) | 110 (6.23%) | 446 (25.3%) |
| | Total | | 3,349 | 258 (7.7%) | 2,030 (60.6%) | 351 (10.5%) | 710 (21.2%) |
| UKBB | 1 | 5 | 552 | 23 (4.17%) | 34 (6.16%) | 470 (85.1%) | 25 (4.53%) |
| | 2 | 10 | 252 | 0 (0%) | 166 (65.9%) | 76 (30.2%) | 10 (3.97%) |
| | Total | | 804 | 23 (2.86%) | 200 (24.9%) | 546 (67.9%) | 35 (4.35%) |

Relative to the total number of CFIs in each database, 0.588% of CFI images in the Rotterdam Study evaluation set were identified by the audit as being misadjudicated, with 0.164% for AREDS and 0.259% for the UK Biobank. For UK Biobank, these were identified as cases where the laterality (left or right eye) of the images was incorrectly recorded, rather than the patient identifier. A further 0.615% of images in the Rotterdam Study, 0.224% of images in AREDS and 6.15% of images in the UKBB were identified during the audit as showing no recognizable anatomy (*Anatomy not Visible*).

## Patient verification

We present verification performance results for several application scenarios and sub-scenarios (see *Patient verification evaluation* in the Methods section).

Table 2 presents the AUROC, false positives per 1000 cases (FP/1000), and false negatives per 1000 cases (FN/1000) for each scenario at a fixed operating point tuned in the held-out validation set. FN/1000 is also reported for different levels of FN/1000 in the evaluation set. The mean number of reference images available to the model is reported, since it is a relevant variable that changes with the sub-scenarios evaluated. All scenarios, except for the "Near-duplicates" scenario involve the removal of near-duplicate images.

*Table 2. Evaluation of patient verification scenarios in a held-out set of identities from the Rotterdam Study. In this application, the model decides whether to flag an image as an identity mis-assignment given the image itself and a reference set of images from the corresponding patient. N indicates the number of cases evaluated. Mean ref. images is the mean number of reference images available from the patient under the scenario. Threshold @ 20 FP/1000 indicates that the metrics were measured using*

*a fixed decision threshold that was calibrated on the held-out validation set for 15 false positives per 1000 images. FN/1000 at fixed FP/1000 indicates the false negative rates measured at a given false positive rate on the benchmark set.*

| **Scenario** | **N** | **Mean ref. images** | **Threshold @ 15 FP/1000 (val)** | | | **FN/1000 at fixed FP/1000** | | |
|---|---|---|---|---|---|---|---|---|
| | | | **AUROC** | **FP/1000** | **FN/1000** | **FP=3** | **FP=5** | **FP=10** |
| Retrospective-only verification | | | | | | | | |
| Base (near-duplicates included) | 78518 | 6.1 | 0.9998 | 3 | 3.5 | 3.5 | 1.1 | 0.1 |
| Base (near-duplicates removed) | 48470 | 3.9 | 0.9998 | 2.7 | 2.5 | 2 | 0.4 | 0.1 |
| Same device | 20330 | 1.6 | 0.9996 | 6.1 | 2.8 | 12.9 | 4.8 | 0.5 |
| Disc-centred images (F1) | 13150 | 2.1 | 1 | 0.2 | 3.3 | 0.2 | 0.2 | 0 |
| Macula-centred images (F2) | 30816 | 2.5 | 0.9998 | 4.1 | 2.7 | 5.2 | 1.5 | 0.3 |
| Image quality Q1+Q2 (best) | 20476 | 2.6 | 1 | 0.2 | 2.4 | 0 | 0 | 0 |
| Image quality Q3+Q4 (worst) | 20158 | 2.6 | 0.9996 | 5.6 | 3.9 | 13.7 | 5.8 | 1.3 |
| Retrospective + prospective verification | | | | | | | | |
| Base (near-duplicates included) | 110272 | 13.1 | 0.9999 | 1 | 5.2 | 0.1 | 0 | 0 |
| Base (near-duplicates removed) | 70548 | 7.5 | 0.9998 | 1.6 | 4.5 | 0.9 | 0.1 | 0.1 |
| Same device | 67224 | 1.8 | 0.9999 | 2 | 2.7 | 0.5 | 0.1 | 0 |
| Disc-centred images (F1) | 24422 | 3.1 | 1 | 0.2 | 4 | 0.1 | 0 | 0 |
| Macula-centred images (F2) | 44626 | 4.1 | 0.9998 | 2.5 | 3 | 2 | 0.4 | 0 |
| Image quality Q1+Q2 (best) | 34626 | 4.5 | 1 | 0.1 | 4.9 | 0 | 0 | 0 |
| Image quality Q3+Q4 (worst) | 31934 | 4.7 | 0.9996 | 3.5 | 5.3 | 9.4 | 2 | 0.1 |

Table 3 presents equivalent results from zero-shot external validation on the AREDS and UK Biobank CFI databases. Here, the number of scenarios has been reduced due to some scenarios not being applicable in these datasets. For example, the "Same device" scenario does not apply since both AREDS and UKBB contain a single device.

*Table 3. External evaluation of the system in different patient verification scenarios on the AREDS and UK Biobank CFI databases. In this application, the model decides whether to flag an image as an identity mis-assignment given the image itself and a reference set of images from the corresponding patient. N indicates the number of cases evaluated. Mean ref. images is the mean number of reference images available from the patient under the scenario. Threshold @ 10 FP/1000 indicates that the metrics were measured using a fixed decision threshold that was calibrated on the held-out validation set for 10 false positives per 1000 images. FN/1000 at fixed FP/1000 indicates the false negatives rates measured at a given false positive rate on the benchmark set.*

| **Scenario** | **N** | | **Threshold @ 10 FP/1000 (val)** | **FN/1000 at fixed FP/1000** |
|---|---|---|---|---|

| | | Mean ref. images | AUROC | FP/1000 | FN/1000 | FP=3 | FP=5 | FP=10 |
|---|---|---|---|---|---|---|---|---|
| **UKBB** | | | | | | | | |
| Retrospective-only verification | | | | | | | | |
| Base (no duplicates) | 7742 | 1 | 0.9997 | 13.9 | 0.8 | 16.5 | 9.3 | 1 |
| Image quality Q1+Q2 (best) | 3012 | 1 | 1 | 0 | 0 | 0 | 0 | 0 |
| Image quality Q3+Q4 (worst) | 2630 | 1 | 0.9993 | 35 | 0.8 | 89 | 34.2 | 11.4 |
| Retrospective + prospective verification | | | | | | | | |
| Base (no duplicates) | 15484 | 1 | 0.9996 | 13.9 | 0.8 | 18.9 | 9.7 | 1.3 |
| Image quality Q1+Q2 (best) | 6024 | 1 | 1 | 0 | 0.7 | 0 | 0 | 0 |
| Image quality Q3+Q4 (worst) | 5260 | 1 | 0.9992 | 35 | 1.5 | 84.8 | 29.3 | 14.8 |
| **AREDS** | | | | | | | | |
| Retrospective-only verification | | | | | | | | |
| Base (near-duplicates included) | 257302 | 11.9 | 0.9998 | 2.7 | 4.1 | 3 | 0.5 | 0.1 |
| Base (near-duplicates removed) | 129042 | 6.4 | 0.9998 | 3.7 | 2.9 | 5.3 | 1.1 | 0.1 |
| Disc-centred images (F1) | 12942 | 3.3 | 0.9999 | 0.6 | 0.6 | 0 | 0 | 0 |
| Macula-centred images (F2) | 116100 | 4.5 | 0.9997 | 4.9 | 2.3 | 7.6 | 2.3 | 0.2 |
| Image quality Q1+Q2 (best) | 60378 | 4.7 | 1 | 0.6 | 1.7 | 0 | 0 | 0 |
| Image quality Q3+Q4 (worst) | 56888 | 3.8 | 0.9987 | 18.1 | 3.4 | 116.7 | 56.9 | 17.6 |
| Retrospective + prospective verification | | | | | | | | |
| Base (near-duplicates included) | 297328 | 22.2 | 0.9999 | 0.5 | 5.5 | 0 | 0 | 0 |
| Base (near-duplicates removed) | 148382 | 11.4 | 0.9998 | 2 | 4.4 | 1.7 | 0.3 | 0 |
| Disc-centred images (F1) | 15318 | 5.6 | 0.9999 | 0.4 | 1.4 | 0 | 0 | 0 |
| Macula-centred images (F2) | 133056 | 7.8 | 0.9998 | 2.8 | 2.7 | 2 | 0.3 | 0 |
| Image quality Q1+Q2 (best) | 75048 | 7.9 | 1 | 0.3 | 2.9 | 0 | 0 | 0 |
| Image quality Q3+Q4 (worst) | 69544 | 6.2 | 0.9997 | 5.3 | 4.7 | 13.3 | 5.4 | 0.9 |

## Image-based retrieval of the correct patient identity

We measured retrieval performance for several application scenarios and sub-scenarios (see *Patient identity retrieval evaluation* in the Methods section). Table 4 presents the retrieval metrics – Recall@1, Recall@5 and mAP – for the scenario in which the model has access only to prior imaging for each identity (Retrospective-only scenario). Results for the Retrospective + Prospective scenario are included in Appendix D.

*Table 4. Results of the retrospective patient identity retrieval evaluation on the Rotterdam Study, UKBB, and AREDS CFI databases. In this scenario that system had access only to imaging acquired previous to the query image. N queries indicates the number of images queried, Mean comparison identities is the average number of identities in the gallery, across all queries.*

| Scenario | N queries | Mean comparison identities | Mean images / identity | Recall@1 | Recall@5 | mAP |
|---|---|---|---|---|---|---|
| **Rotterdam Study** | | | | | | |
| Base (near-duplicates included) | 3360 | 4482 | 4.7 | 0.997 | 0.998 | 0.998 |
| Base (near-duplicates removed) | 3271 | 4435.8 | 3.3 | 0.997 | 0.997 | 0.997 |
| Same device | 3058 | 3082 | 1.9 | 0.993 | 0.994 | 0.994 |
| Disc-centred images (F1) | 1540 | 3186.7 | 2.5 | 1 | 1 | 1 |
| Macula-centred images (F2) | 3170 | 4307.5 | 2.6 | 0.995 | 0.995 | 0.995 |
| Image quality Q1+Q2 (best) | 2015 | 3532.6 | 2.3 | 1 | 1 | 1 |
| Image quality Q3+Q4 (worst) | 2009 | 3706 | 2.3 | 0.992 | 0.995 | 0.993 |
| **UK Biobank** | | | | | | |
| Base (near-duplicates included) | 1971 | 4339 | 1 | 0.975 | 0.989 | 0.981 |
| Base (near-duplicates removed) | 1976 | 4339 | 1 | 0.972 | 0.987 | 0.979 |
| Image quality Q1+Q2 (best) | 779 | 2472 | 1 | 0.999 | 1 | 0.999 |
| Image quality Q3+Q4 (worst) | 688 | 1867 | 1 | 0.935 | 0.974 | 0.951 |
| **AREDS** | | | | | | |
| Base (near-duplicates included) | 7084 | 8511.3 | 8.8 | 0.983 | 0.989 | 0.986 |
| Base (near-duplicates removed) | 7128 | 8509.8 | 4.7 | 0.976 | 0.986 | 0.98 |
| Disc-centred images (F1) | 993 | 1189.8 | 3.3 | 0.998 | 0.998 | 0.998 |
| Macula-centred images (F2) | 7123 | 8515.6 | 3.7 | 0.985 | 0.991 | 0.988 |
| Image quality Q1+Q2 (best) | 4797 | 6676.9 | 3.1 | 0.989 | 0.992 | 0.991 |
| Image quality Q3+Q4 (worst) | 5240 | 6787.9 | 3 | 0.943 | 0.962 | 0.952 |

# Discussion

Our results demonstrate that an AI model trained for patient re-identification on the large, diverse, population-scale color fundus imaging dataset from the Rotterdam Study can accurately detect patient assignment errors and retrieve the correct patient identity from a database of thousands of individuals. The model remained robust across imaging devices and a wide range of patient ages.

The expert verification of images flagged by the model as having potentially incorrect identities ( Table 1) revealed that large-scale population databases such as the Rotterdam Study, UK Biobank and AREDS do contain a small proportion of mis-adjudicated images. We estimated this proportion to be up to 0.6% in the Rotterdam Study and 0.2% in both AREDS and UKBB. The identity audit served as a case study demonstrating the application of our system, confirming that the identity embedding distances between target and reference images provided a clear signal for outlier detection. It further highlighted image quality and differences in retinal region – particularly prevalent in AREDS – as factors that can challenge the model and, in some cases, make adjudication impossible even for a human grader.

Evaluation on the diverse Rotterdam Study validation set, comprising images acquired with multiple image devices, varying image quality, and both macula and optic disc-centered fields (

Table 2) demonstrated strong performance across all evaluation scenarios. AUROC provided limited discrimination between scenarios because performance approached saturation on most benchmarks. As expected, in the fixed threshold evaluation (Threshold @ 15 FP/1000), the observed false positives rates (FP/1000) were lower than the calibrated 15 FP/1000 because the calibration set was not audited and corrected. FP/1000 and FN/1000 varied substantially across scenarios, likely due domain shifts in database characteristics due to scenario assumptions.

Retrospective-only verification, as expected, was more challenging than when prospective imaging was also available to the system. Removal of near-duplicate images (same retinal region, visit and imaging device) had no discernible effect on performance. The likely reason is that in the RS the same retinal region is only imaged more than once during the same visit when the quality of previous photographs is not satisfactory. Limiting query and reference images to the same device (Same device scenario) resulted in only a modest change in performance in both scenarios, despite a much lower number of reference images – as low as 1.6 for the retrospective scenario. Stratification by retinal region (Disc-centred and Macula-centred scenarios) revealed better performance for images centered on the optic disc, with as low as 0.2 FP/1000 and 3.3 FN/1000 for the retrospective scenario. This difference is likely due a better visibility of the major retinal vessels surrounding the optic disc. In many macula-centered images from RS, the field of view includes only a partial view of the optic disc and major retinal vascular arcades, which limits the anatomy available for identity matching. Image quality played an important role on model performance, with the evaluation on images above and below the $50^{th}$ percentile of quality having the highest and lowest scores across all scenarios respectively.

Zero-shot external evaluation on UKBB and AREDS (

Table 3) showed similar strong performance, indicating that the model generalizes well across datasets acquired under different imaging conditions. Performance on the UKBB was high despite this database having only one reference image per query. Performance on AREDS was on par, despite this dataset containing non-standard imaging regions. In both cases, the model achieved perfect or near-perfect performance when restricting the database to disc-centered images, or to the top 50% of images by quality. In both databases, the difference in performance between image quality levels was larger than for the RS. Reasons for this may include domain differences that affect transfer performance differently across quality levels, and potentially a different, less dispersed distribution of image quality levels in these databases.

The retrieval evaluation tested the ability of the model to find the correct patient identity given an image. Retrieval performance was consistently high across datasets and evaluation scenarios. Recall@1 was consistently above 0.97 in the primary retrospective-only scenario after duplicate removal, indicating that the model retrieved the correct identity in more than 97% of cases. In the Rotterdam Study the system ranked the correct identity first in 99.7% of cases in the RS, retrieving the correct individual from a database of 4,436 identities on average, by comparing with 3.3 reference images per identity. The correct identity was ranked amongst the top 5 retrieval results (Recall@5) in at least 98.6% of cases for all databases. Mean average precision (mAP) had a consistent behavior. The size of the gallery – measured as the mean number of comparison identities (Table 4) did not appear to drive performance.

Sub-scenario results were consistent with the identity verification evaluation. In the RS, reducing the gallery to images from the same device as the query resulted in a modest drop in performance in the Retrospective scenario but not in the Retrospective + Prospective scenario, which saw an improvement. This can be attributed to a reduction in the number of matching images of the same retinal region after removal of other-device images. This is likely to affect retrospective queries asymmetrically because they are less likely to have an image with matching field and device in the gallery when compared to the same *Retrospective + Prospective* query. Retrieval performance was the highest when filtering to F1 images only, and for the top 50% of images by quality, across databases. For images centered on the optic disc, Recall@1 was 100% for the RS and as high as 99.8% for AREDS.

Our system achieved retrieval scores substantially higher than those reported in previous work. Nebbia et al.[42] report image-level and patient-level recall scores for a foundation model fine-tuned for CFI retrieval. In their image-level evaluation, equivalent to our evaluation setup, the authors report a Recall@1 of 82.3, internally evaluated on a held-out test set, using a dataset not available publicly. The size of the evaluation gallery is not reported explicitly but is mentioned to be 20% of a 901-patient longitudinal cohort, which is equivalent to 180 patients. The dataset is reported to have 21.7 images per patient on average. There are several likely reasons for the superior performance of our system on larger evaluation sets with less images per patient – which represent more challenging scenarios. Our development set is unique in its diversity – the same eyes were imaged by multiple devices – and long follow up of up to 32 years. The variability in image conditions and the longitudinal differences present in the dataset due to retinal aging likely acted as natural regularizers for our model. Furthermore, we applied best practices from computer vision person re-identification in the development of our model architecture and training pipeline, with strong augmentations, hard triplet mining, a relatively high resolution of 384x384 and a composite loss.

In accordance with the quantitative results, inspection of failure cases (Supplementary Appendices E and F) revealed that many algorithm mistakes can be explained by imaging factors and anatomical similarity. Most patient verification false positives relate to query images with severe occlusion, blurriness; or an imaging region different from the reference set. Most false negatives correspond to query images with similar vascular patterns as those in the reference images. Similarly, retrieval failures are often for query images with poor quality, severe occlusion of the anatomy, or a non-standard imaging region without visibility of the optic disc. In these cases, the top matching image generally has similar characteristics.

Our preliminary analysis of the predictive power of identity embeddings towards imaging, demographic and disease variables (Appendix A) revealed that the identity embeddings encode these variables weakly. Ethnicity stood out as the most predictable variable by AUC. However, the RS is highly imbalanced with respect to ethnicity, with around 97% of participants being European, making ethnicity alone unlikely to explain the model's re-identification performance. Nevertheless, previous work has shown that retinal pigmentation, which is associated to ethnicity, can be measured reproducibly from CFIs[43]. We hypothesize that our identity encoder may have learned to use retinal pigmentation as a stable cue for re-identification. The imaging device was more difficult to predict in comparison, despite its obvious effect on overall image appearance. Age and sex were also less predictable, with AUC scores considerably lower than those reported in previous work for dedicated models[44]. Among the disease variables, age-related macular degeneration showed the strongest association, potentially due to lesions providing a strong re-identification cue. Overall, these results are consistent with an encoder that relies primarily on anatomical pattern matching, but is not completely agnostic to imaging, demographics, and disease-related factors.

In summary, our results provide the first evidence that a retinal biometrics system for patient identity verification and retrieval can achieve very high performance on diverse longitudinal datasets spanning multiple image devices, participant demographics, and image acquisition conditions representative of both research and clinical settings. In ophthalmology, where clinical care is highly image-driven and longitudinal, we consider that AI-assisted verification and retrieval workflows have the potential to drastically improve the time-effectiveness of identity auditing in clinical and research settings, potentially making new workflows cost-effective; and enhancing the integrity of longitudinal clinical records and research databases. More broadly, the performance and generalizability of our system motivate further research on CFI-based retinal biometrics for clinical deployment, and the extension to other ophthalmic modalities.

Despite the strong performance of our system, our analysis also identified a key limitation when the anatomy – particularly vascular patterns – are not consistently visible in the images. In applications with high proportion of such low quality images, we recommend the deployment of our algorithm in conjunction with a suitable quality filtering algorithm[41,45]. Furthermore, we do not recommend its application as a fully autonomous correction system without rigorous validation on the target context.

## A word on privacy

Our work shows that AI systems can use the retinal anatomy as a powerful biometric identifier. Throughout this paper, "biometrics" and related terms are used in a technical rather than legal sense, without making any determination as to whether retinal images or derived representations constitute biometric data under applicable law. While our work opens the door for new clinical applications, this technology could also be used maliciously to de-identify clinical and research databases that are otherwise considered anonymous. This scenario would require an attacker to be in possession of a second reference database with retinal imaging linked to patient identities. A good identity encoder would allow the attacker to cross-reference images across databases, thereby de-anonymizing patient information. We consider this a highly unlikely scenario due to the current lack of accessible reference databases containing retinal imaging linked to sensitive patient data. Such databases are most often limited to clinical institutions and are protected from public sharing by privacy laws. However, we acknowledge that data leaks of clinical and research data are an ever-present risk[46].

Furthermore, in recent years, policy discussions around the sharing of personal (medical) data have often centered around the risk of de-identification, with uncertainty about the status of retinal images as biometric identifiers[47–49]. We note that throughout this paper, "biometrics" and related terms are used in a technical rather than legal sense, without making any determination as to whether retinal images or derived representations constitute biometric data under applicable law. However, we acknowledge that the public availability of algorithms for patient identity matching has the potential to alter policy decisions around the open sharing of ophthalmic imaging, potentially curtailing research progress. For that reason, instead of being publicly available, our model will be made available to verified parties for clinical and research purposes under an End User License Agreement (EULA).

# Author Contributions

JDVQ led the conceptualization and planning of the study, designed and developed the proposed system, performed the primary evaluation on the Rotterdam Study and AREDS, and drafted the manuscript. DB contributed with the execution of the identity audit and evaluation on the UKBB and provided feedback on the manuscript. JV and BL contributed with data management related to RS and AREDS, and provided feedback on the manuscript. SB conceived, acquired funding for, and led the VascX Research Consortium. CCWK secured funding for VascX and the RS, supervised ophthalmic data collection within RS, and provided feedback on the manuscript. MT provided feedback on the manuscript. All authors reviewed and approved the final manuscript.

# Acknowledgments & Funding

The authors are grateful to the study participants, the staff from the Rotterdam Study and the participating general practitioners and pharmacists. The authors acknowledge the staff of the EyeNED Reading Center for their contributions.

Funding support for the Age-Related Eye Disease Study (AREDS) Database used in this study was provided by the National Eye Institute (N01-EY-0-2127). We thank the AREDS participants and the AREDS Research Group for their valuable contributions to this research. This research has been conducted using the UK Biobank Resource under Application Number 90947.

This work was funded by the Swiss National Science Foundation grant no. CRSII5 209510.394. The funder played no role in study design, data collection, analysis and interpretation of data, or the writing of this manuscript. The Rotterdam Study is funded by Erasmus Medical Center and Erasmus University, Rotterdam, Netherlands Organization for the Health Research and Development (ZonMw), the Research Institute for Diseases in the Elderly (RIDE), the Ministry of Education, Culture and Science, the Ministry for Health, Welfare and Sports, the European Commission (DG XII), and the Municipality of Rotterdam.

## Competing Interests

The author(s) declare no competing interests.

## Data Availability

Data for the primary Rotterdam Study dataset can be obtained upon request. Requests should be directed towards the management team of the Rotterdam Study (datamanagement.ergo@erasmusmc.nl), which has a protocol for approving data requests. Because of restrictions based on privacy regulations and informed consent of the participants, data cannot be made freely available in a public repository. The Rotterdam Study has been approved by the Medical Ethics Committee of the Erasmus MC (registration number MEC 02.1015) and by the Dutch Ministry of Health, Welfare and Sport (Population Screening Act WBO, license number 1071272-159521-PG). The Rotterdam Study has been entered into the Netherlands National Trial Register (NTR) and into the WHO International Clinical Trials Registry Platform (ICTRP) under shared catalogue number NTR6831. All participants provided written informed consent to participate in the study and to have their information obtained from treating physicians.

UK Biobank data are not publicly available due to participant privacy and confidentiality requirements. However, the data are accessible to eligible researchers upon application and approval by UK Biobank. Researchers interested in accessing these data can apply through the UK Biobank Access Management System. The present study was conducted under UK Biobank application number 90947.

The Age-Related Eye Disease Study (AREDS) Database used in this study was obtained through the database of Genotypes and Phenotypes (dbGaP), accession number phs000001.v3.p1. Access to individual-level clinical data and retinal fundus photographs is subject to approval through the dbGaP controlled-access application process. Researchers may request access in accordance with NIH data access policies.

# Appendix A. Identity embeddings have limited predictive power towards demographic variables

Finally, we investigated whether the learned identity embeddings encoded information beyond patient-eye identity. Supplementary Table 1 presents the performance evaluation of logistic regression models trained to predict demographic and disease variables from the identity embeddings of our model, in the Rotterdam Study evaluation set.

*Supplementary Table 1. Evaluation of CFI identity embeddings for predicting imaging, demographic and disease targets. Results suggest limited discriminative power of identity embeddings towards prediction of these variables.*

| Target | N identities | N train | N test | N classes | Test AUC |
|---|---|---|---|---|---|
| Ethnicity | 5,566 | 3,617 | 1,392 | 4 | 0.933 |
| Imaging device | 6,243 | 4,057 | 1,561 | 11 | 0.775 |
| Patient age (decade) | 6,242 | 4,056 | 1,561 | 6 | 0.751 |
| AMD | 4,527 | 2,942 | 1,132 | 2 | 0.736 |
| Glaucoma | 3,901 | 2,535 | 976 | 2 | 0.695 |
| Hypertension | 5,677 | 3,689 | 1,420 | 2 | 0.631 |
| Patient sex | 6,240 | 4,056 | 1,560 | 2 | 0.704 |
| Smoking | 4,865 | 3,161 | 1,217 | 2 | 0.630 |

Supplementary Table 2 presents t-SNE visualizations of the learned identity embeddings, where each dot represents a single Rotterdam Study CFI. The first panel displays a random sample of identities, while the subsequent panels show the same embeddings stratified by imaging device, patient age, sex, ethnicity, hypertension status (yes/no), age-related macular degeneration (yes/no), glaucoma and smoking status.

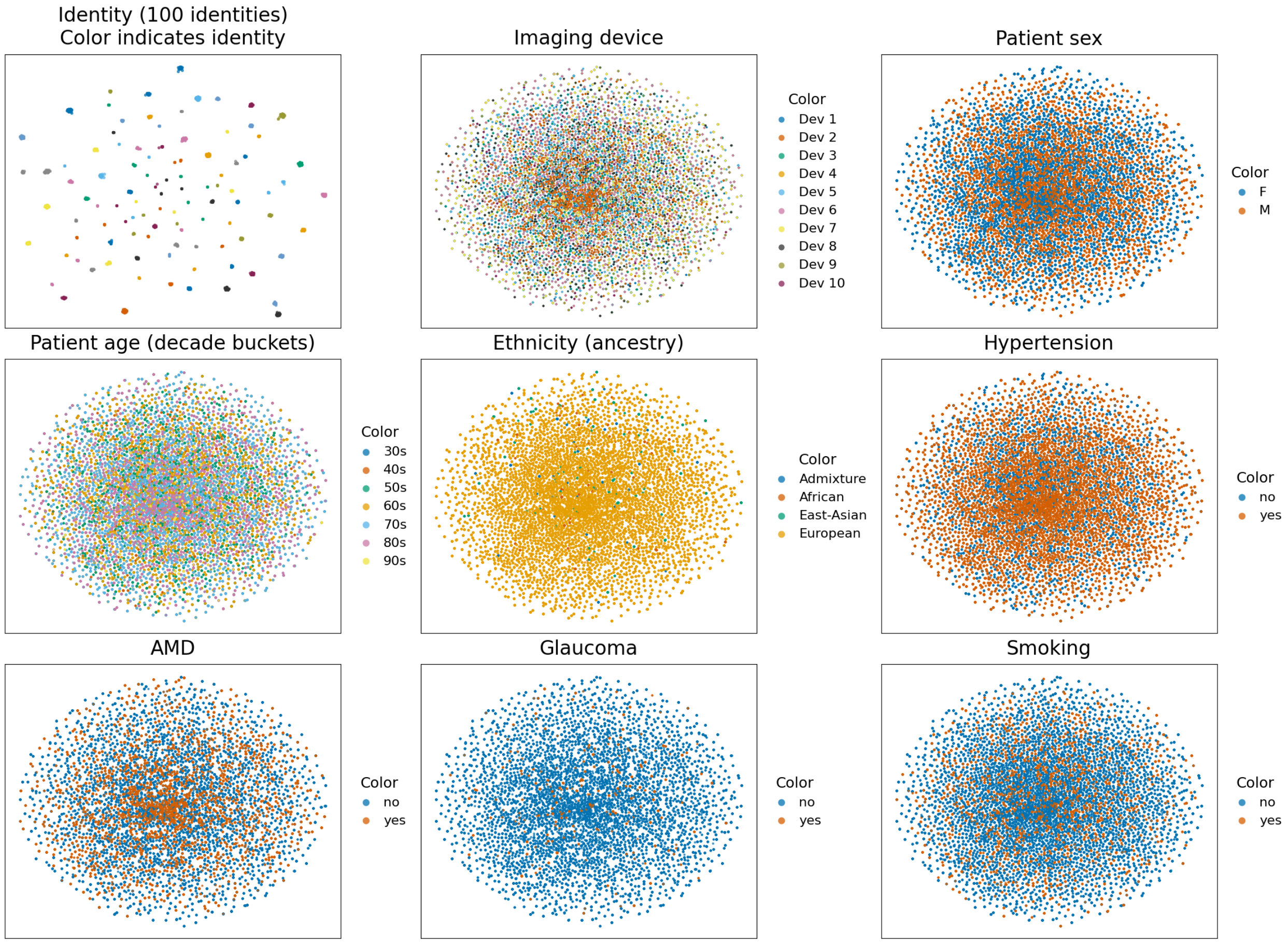

*Supplementary Figure 1. t-SNE visualizations of our model's identity embeddings, color-coded according to patient-eye identity (first cell) and other imaging, demographic and disease variables such as imaging device; patient sex, age, ethnicity and smoking status; and hypertension, age-related macular degeneration, glaucoma and smoking status. The visualization shows that embeddings clearly discriminate between patient-eye identities, but are less discriminative towards imaging, demographic and disease variables.*

# Appendix B. AI model development

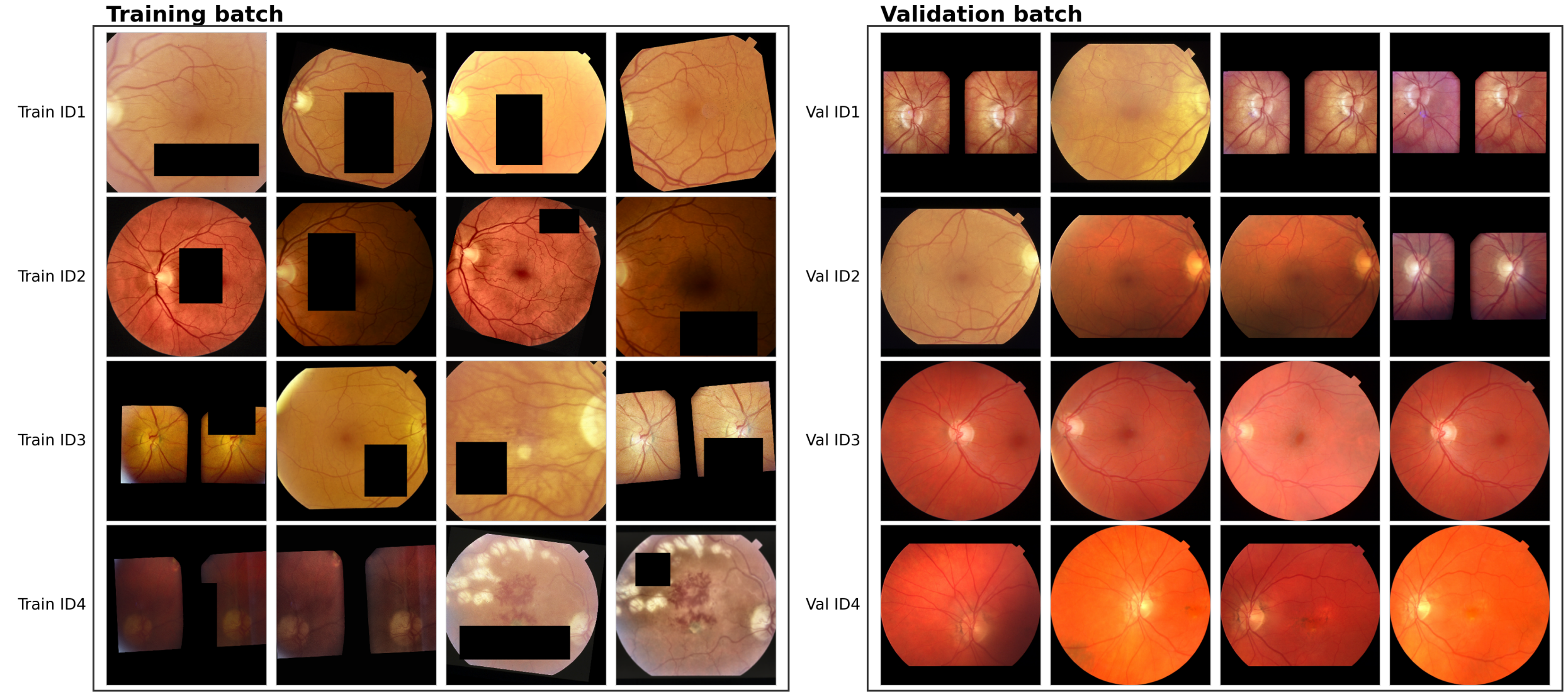

*Supplementary Figure 2. Sample training and validation identities and image samples used in development of the CFI identity encoder. Each row in each batch displays CFIs from the same patient-eye identity. Training samples underwent heavy augmentations while validation samples were kept as-is. Stereoscopic CFIs were included in development.*

# Appendix C. Identity label audit of the evaluation databases.

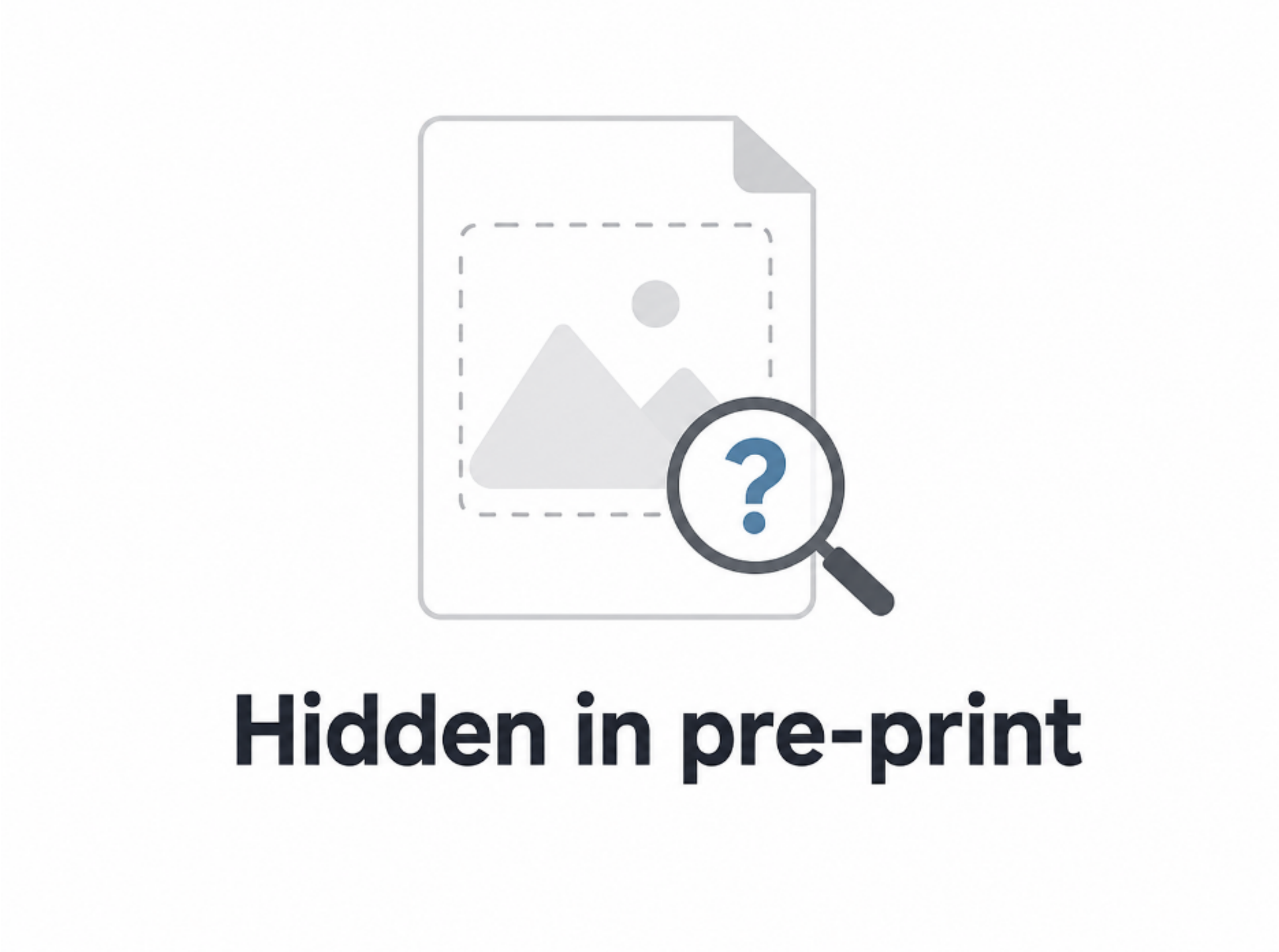

*Supplementary Figure 3. Samples from the identity label audit of the AREDS CFI database. The first three columns display images from the main identity cluster shown to the human expert. The fourth column (Flagged image) shows the image identified by the system as potentially assigned to the wrong identity. The reviewer decided between confirming an incorrect identity, confirming a correct identity, or an undecidable outcome.*

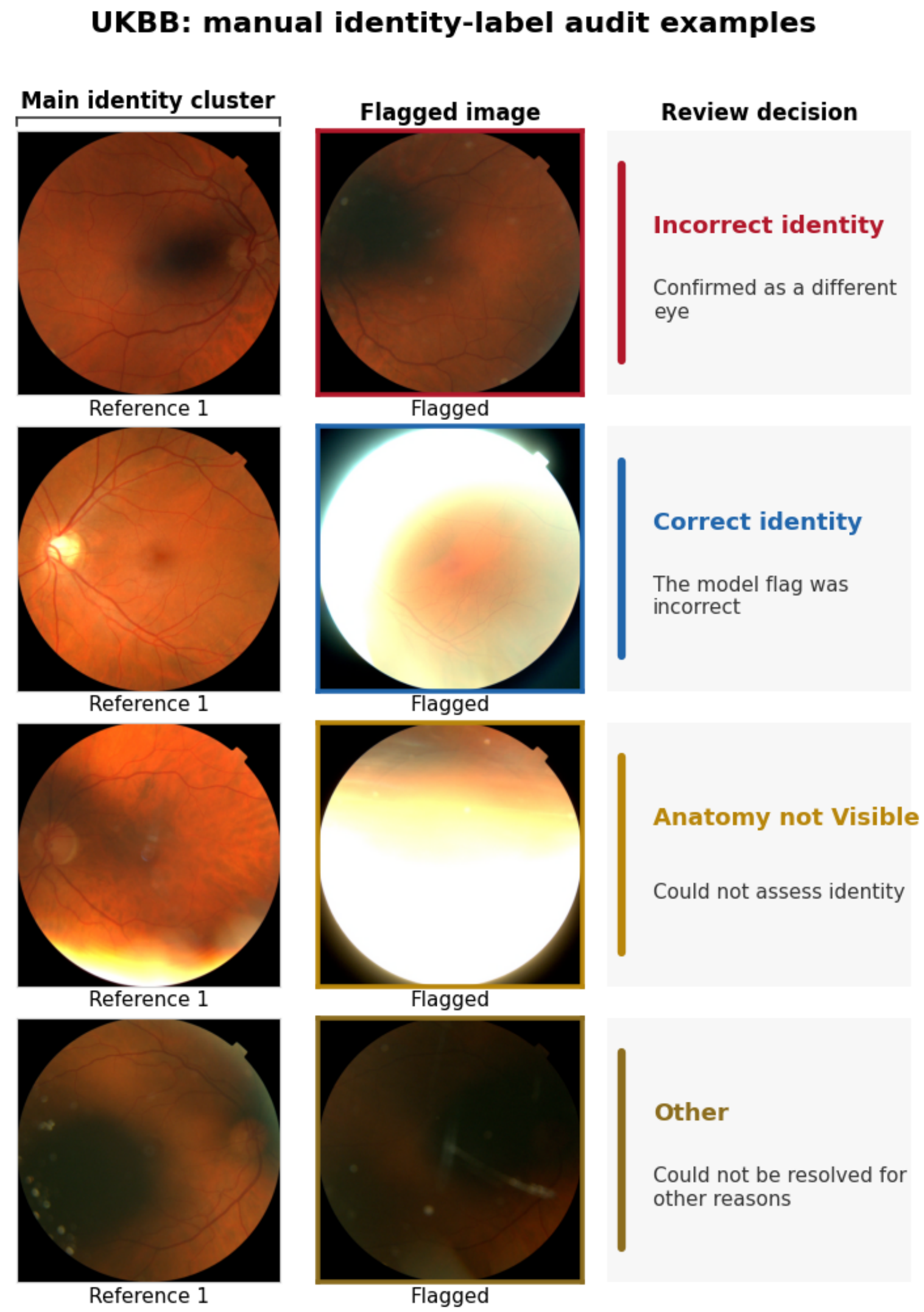

*Supplementary Figure 4. Samples from the identity label audit of the UKBB CFI database. The first column display the representing the main identity cluster shown to the human expert. The second column (Flagged image) shows the image identified by the system as potentially assigned to the wrong identity. The reviewer decided between confirming an incorrect identity, confirming a correct identity, or an undecidable outcome. Images reproduced by kind permission of UK Biobank©.*

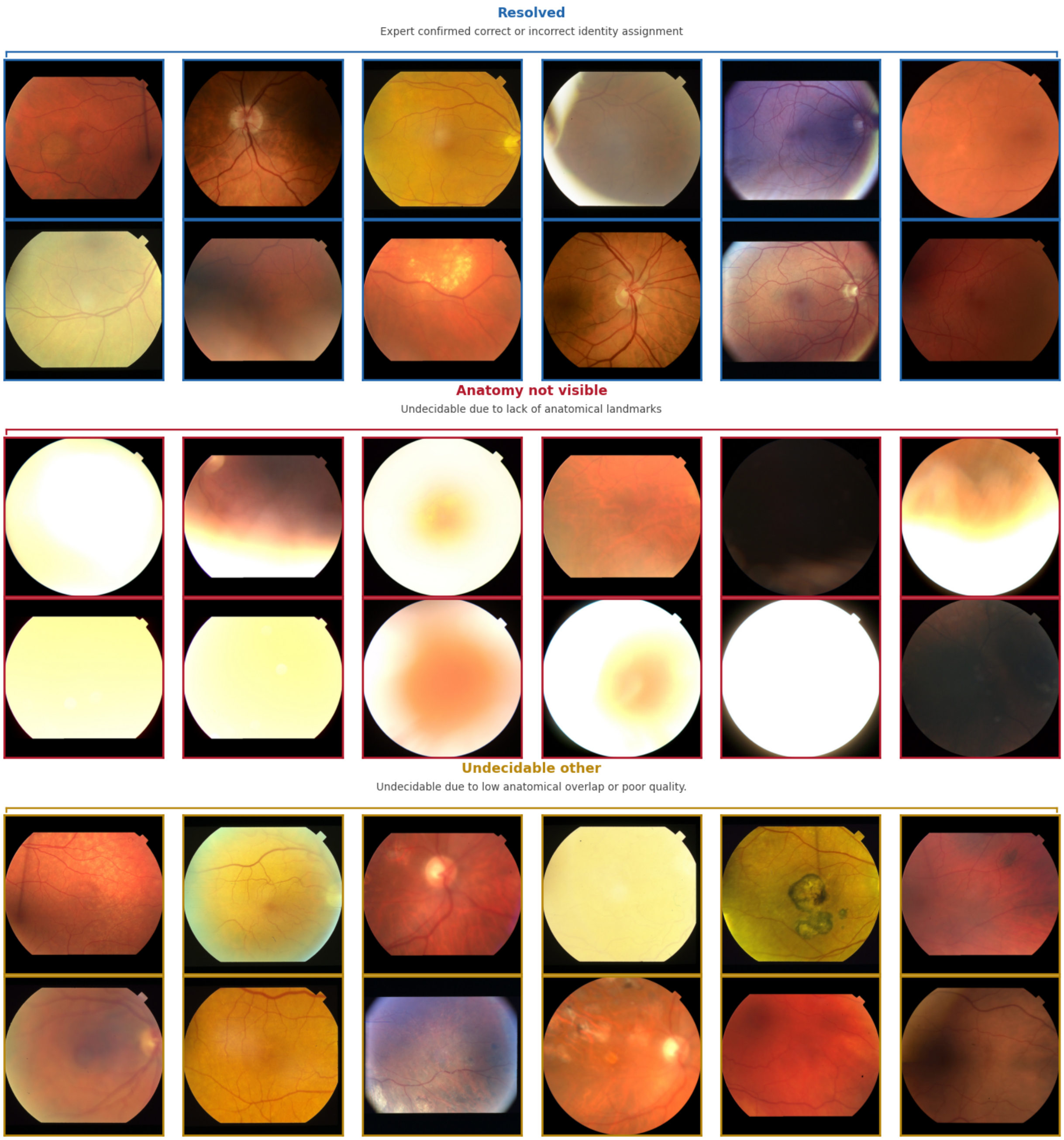

*Supplementary Figure 5. Sample of images flagged during the identity audit of the Rotterdam Study. Most flagged images were of relatively poor quality. Resolved images were confirmed during the expert audit to have a correct or incorrect identity assignment by comparing their vascular patterns to those of the reference images, not seen here. Different region images are undecidable due to imaging a region different from that of the reference images. Low-quality images were those for which the quality was too poor to reach a decision.*

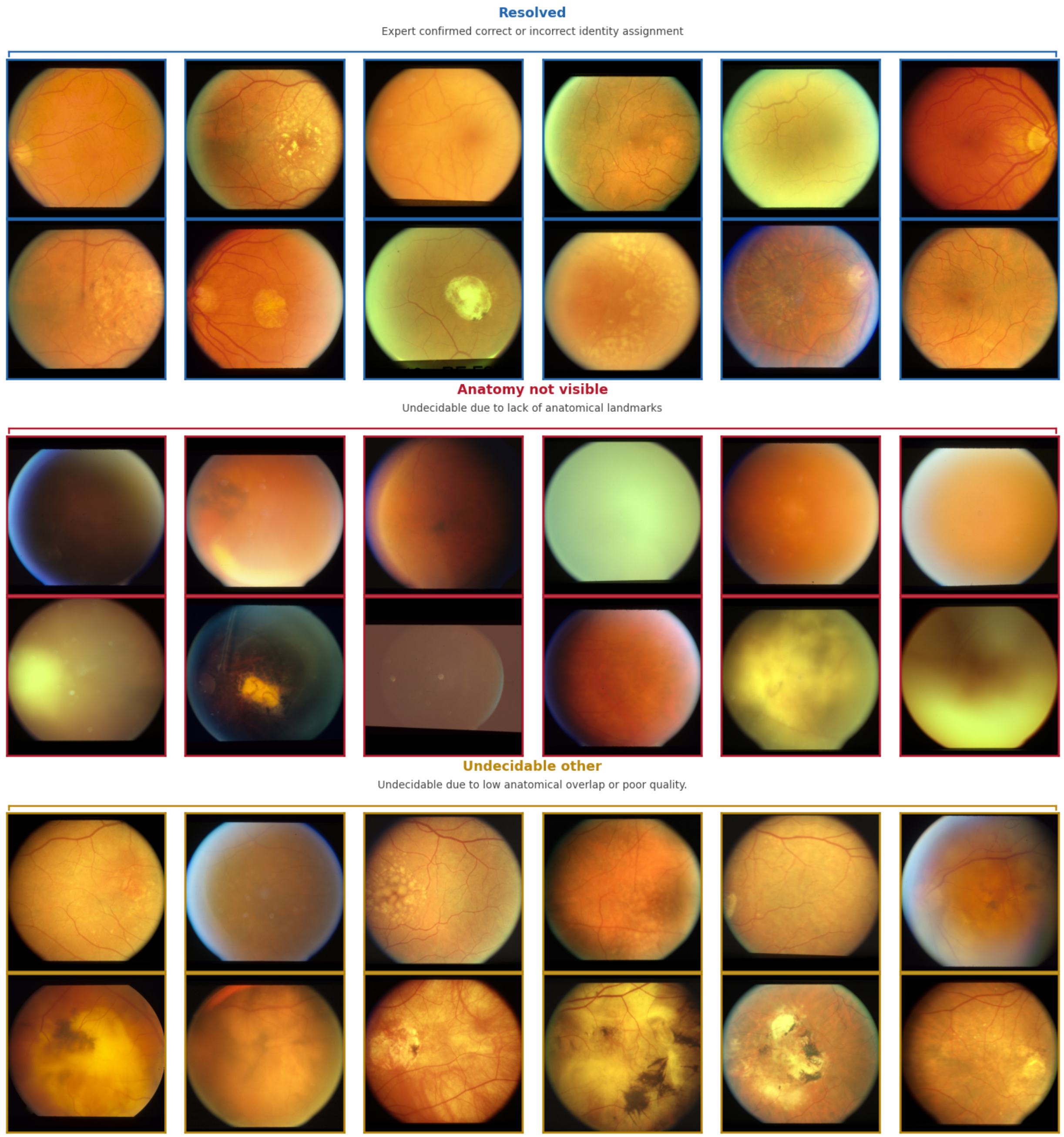

*Supplementary Figure 6. Sample of images flagged during the identity audit of the AREDS CFI database. Most flagged images were of relatively poor quality. Resolved images were confirmed during the expert audit to have a correct or incorrect identity assignment by comparing their vascular patterns to those of the reference images, not seen here. Different region images are undecidable due to imaging a region different from that of the reference images. Low-quality images were those for which the quality was too poor to reach a decision.*

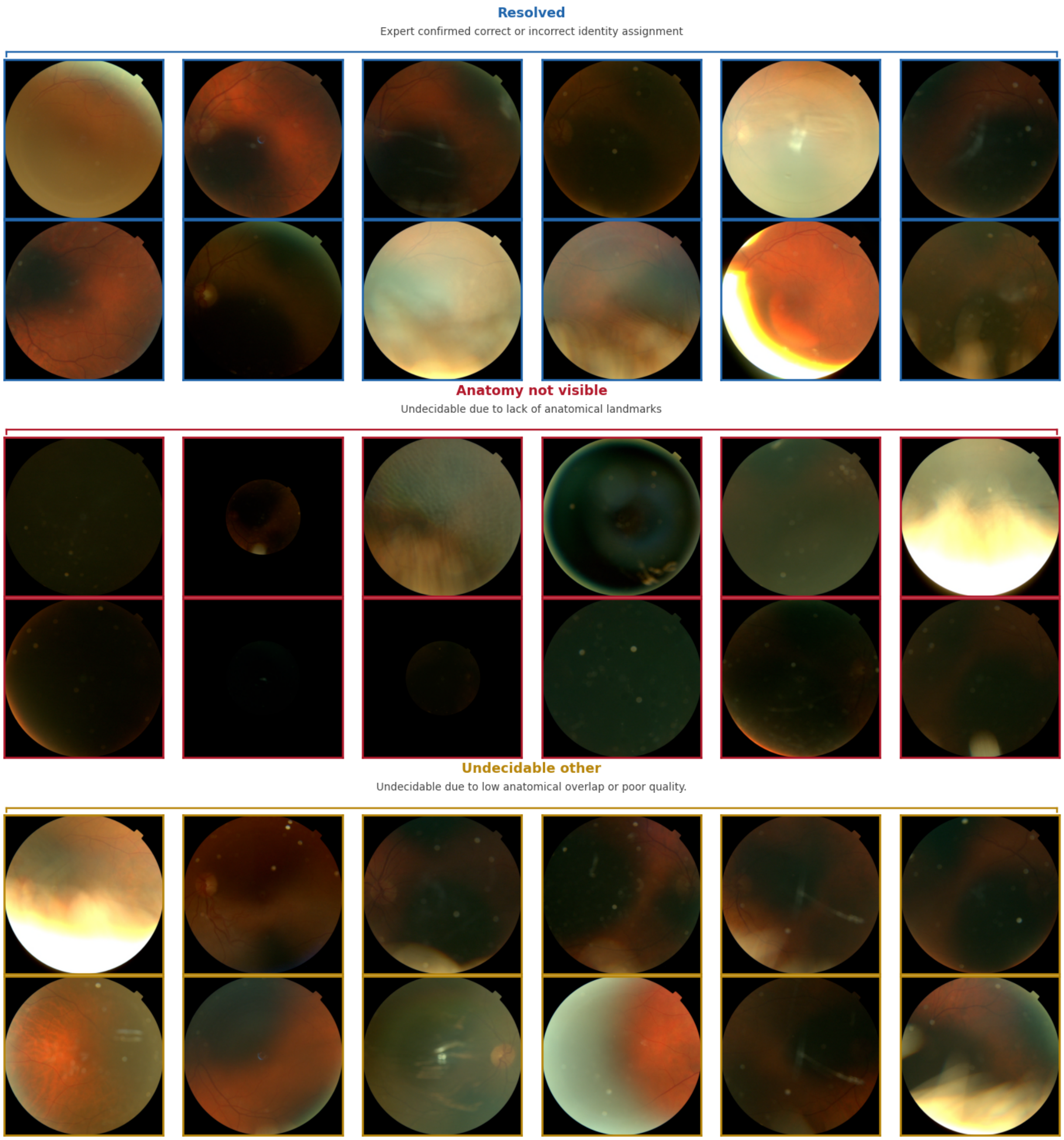

*Supplementary Figure 7. Sample of images flagged during the identity audit of the UK Biobank database. Resolved images were confirmed during the expert audit to have a correct or incorrect identity assignment by comparing their vascular patterns to those of the reference images, not seen here. Different region images are undecidable due to imaging a region different from that of the reference images. Low-quality images were those for which the quality was too poor to reach a decision. Images reproduced by kind permission of UK Biobank©.*

# Appendix D. Complimentary retrieval evaluation

*Supplementary Table 2. Results of the retrospective + prospective patient identity retrieval evaluation on the Rotterdam Study, UKBB, and AREDS CFI databases. In this scenario that system had access to retrospective and prospective imaging relative to the query's acquisition date. N queries indicates the number of images queried, Mean comparison identities is the average number of identities in the gallery, across all queries.*

| Scenario | N queries | Mean comparison identities | Mean images / identity | Recall@1 | Recall@5 | mAP |
|---|---|---|---|---|---|---|
| **Rotterdam Study** | | | | | | |
| Base (near-duplicates included) | 5651 | 6242 | 9 | 0.996 | 0.997 | 0.996 |
| Base (near-duplicates removed) | 5605 | 6242 | 5.8 | 0.994 | 0.996 | 0.995 |
| Same device | 13520 | 2478.6 | 2.3 | 0.997 | 0.998 | 0.998 |
| Disc-centred images (F1) | 3360 | 3847 | 3.3 | 0.999 | 0.999 | 0.999 |
| Macula-centred images (F2) | 5343 | 6238 | 3.7 | 0.994 | 0.996 | 0.995 |
| Image quality Q1+Q2 (best) | 3959 | 5115 | 3.5 | 0.999 | 1 | 1 |
| Image quality Q3+Q4 (worst) | 3859 | 5322 | 3.4 | 0.991 | 0.994 | 0.992 |
| **UK Biobank** | | | | | | |
| Base (near-duplicates included) | 3871 | 4440 | 1.9 | 0.972 | 0.986 | 0.978 |
| Base (near-duplicates removed) | 3871 | 4440 | 1.9 | 0.972 | 0.986 | 0.979 |
| Image quality Q1+Q2 (best) | 1506 | 2649 | 1.6 | 0.999 | 1 | 0.999 |
| Image quality Q3+Q4 (worst) | 1315 | 2841 | 1.5 | 0.929 | 0.96 | 0.944 |
| **AREDS** | | | | | | |
| Base (near-duplicates included) | 8872 | 8881 | 17.6 | 0.99 | 0.996 | 0.993 |
| Base (near-duplicates removed) | 8480 | 8881 | 9.3 | 0.981 | 0.99 | 0.985 |
| Disc-centred images (F1) | 1188 | 1190 | 6.4 | 1 | 1 | 1 |
| Macula-centred images (F2) | 8478 | 8881 | 7.5 | 0.986 | 0.991 | 0.988 |
| Image quality Q1+Q2 (best) | 6467 | 7684 | 5.3 | 0.992 | 0.996 | 0.994 |
| Image quality Q3+Q4 (worst) | 6989 | 8067 | 5.1 | 0.958 | 0.975 | 0.966 |

# Appendix E. Patient verification failure cases

## Retrospective-only scenario

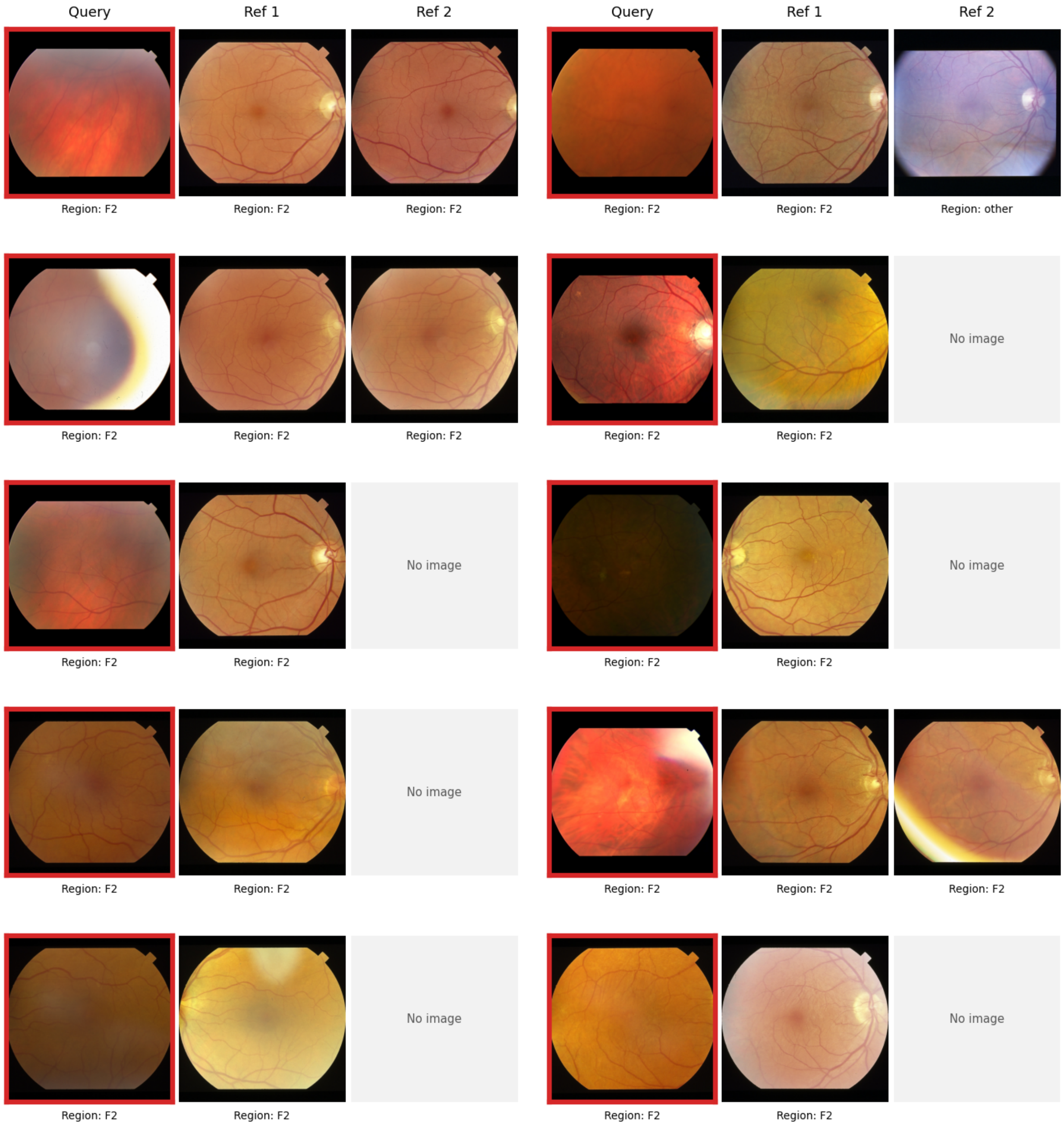

*Supplementary Figure 8. Sample of patient verification false positives produced by our system in the evaluation of the Retrospective-only scenario on the Rotterdam Study database. Each group of three images (Query + Ref 1 + Ref 2) represents the query image and two reference images as presented to the verification system. The three images are assigned to the same patient-eye identity in the reference database, but the verification system flagged an incorrect identity. Reference sets larger than two images were randomly sampled due to space constraints. The vascular patterns partially visible on the images can be used to visually verify that the images correspond to the same eye in most cases. In others, the vascular patterns may be difficult to match.*

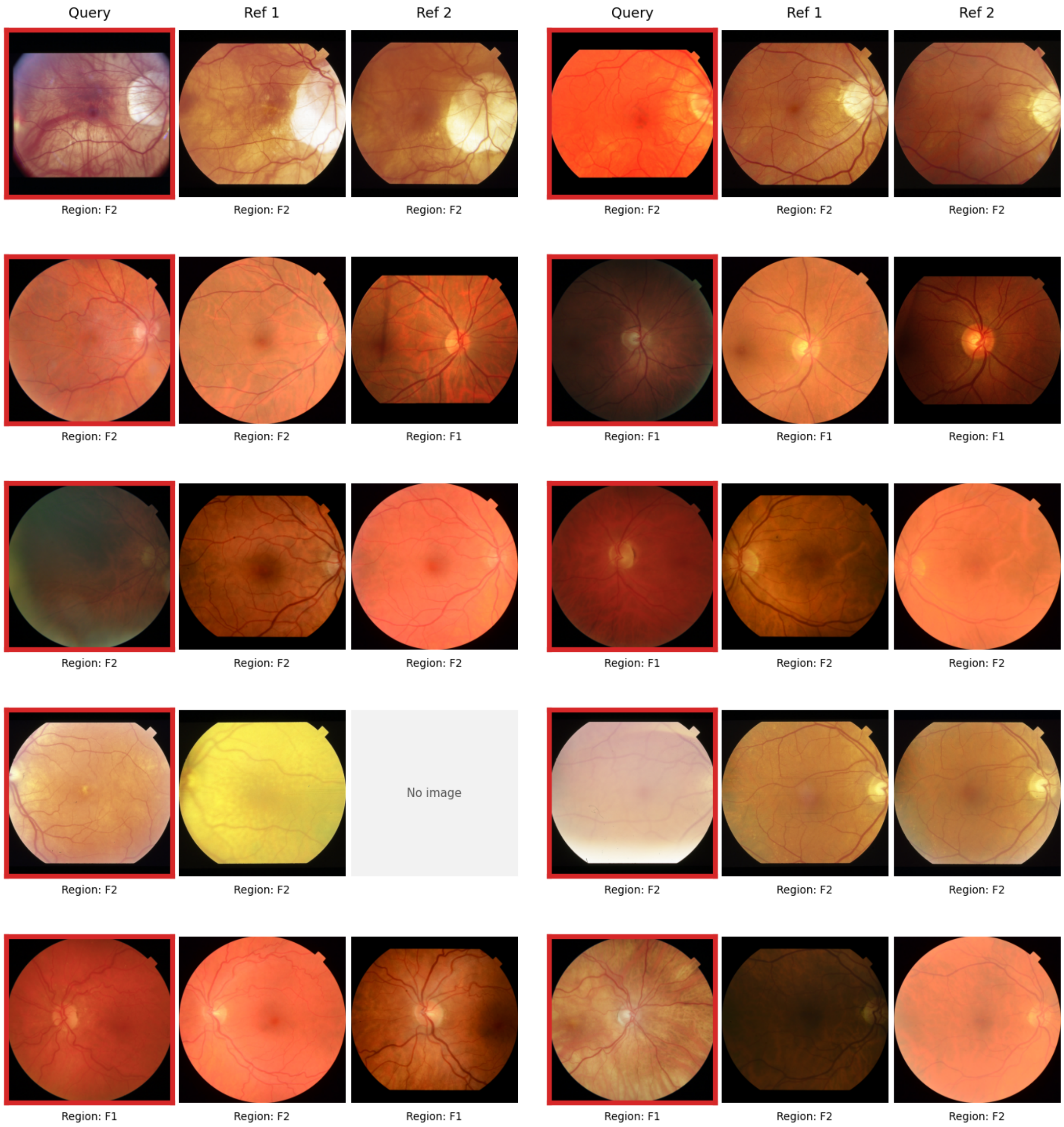

*Supplementary Figure 9. Sample of patient verification false negatives produced by our system in the evaluation of the Retrospective-only scenario on the Rotterdam Study database. Each group of three images presents a simulated incorrect identity assignment consisting in two reference images (Ref 1 and Ref 2) and a query image (marked in red) that was deliberately sampled from a different eye (see Evaluation of patient verification and identity retrieval performance). The model failed to flag an incorrect assignment.*

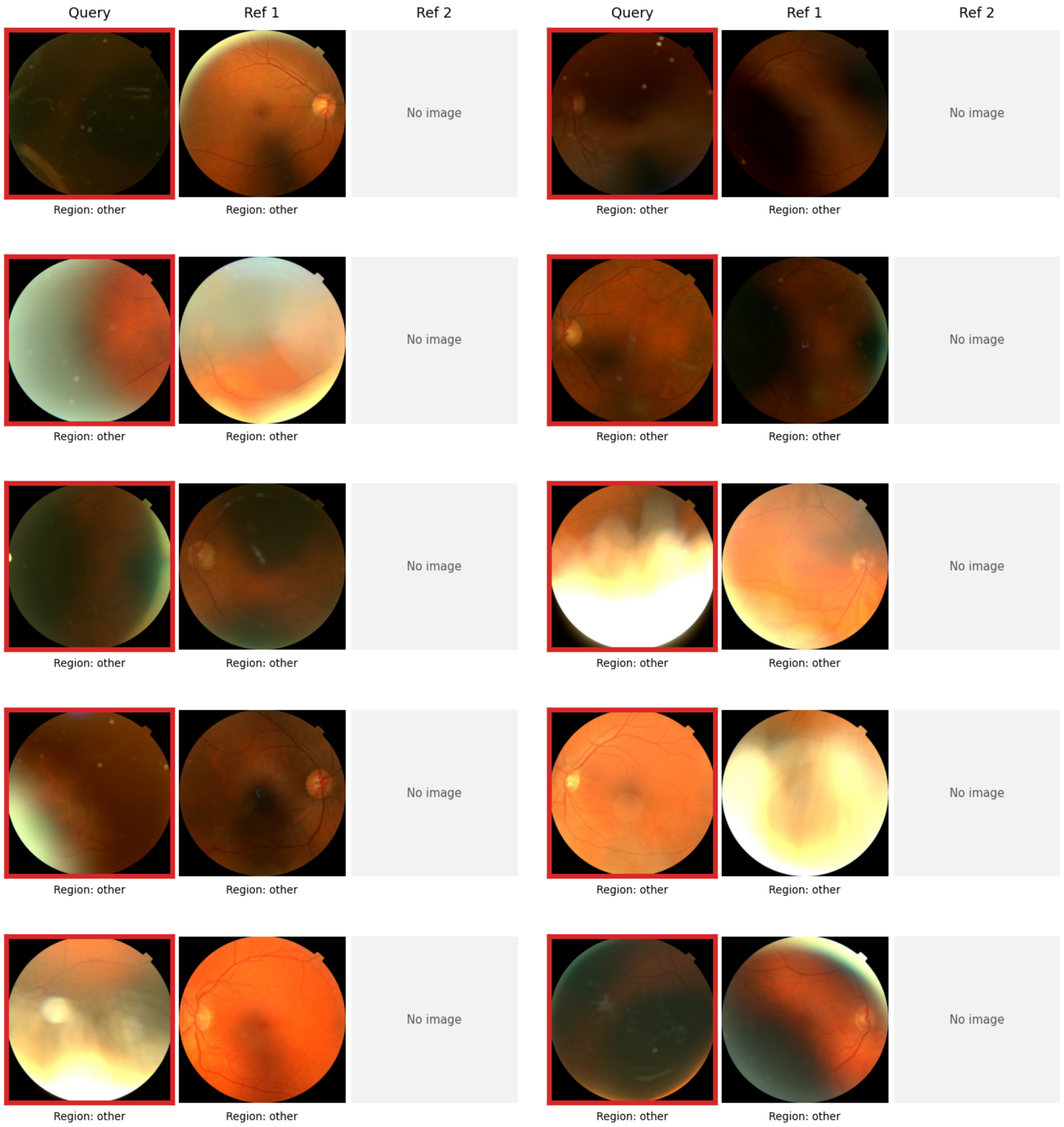

*Supplementary Figure 10. Sample of patient verification false positives produced by our system in the evaluation of the Retrospective-only scenario on the UK Biobank database. Each group of images (Query + Ref 1) represents the query image and reference image as presented to the verification system. The two images are assigned to the same patient-eye identity in the reference database, but the verification system flagged an incorrect identity. The vascular patterns partially visible on the images can be used to visually verify that the images correspond to the same eye in most cases. In others, the vascular patterns may be difficult to match. Images reproduced by kind permission of UK Biobank©.*

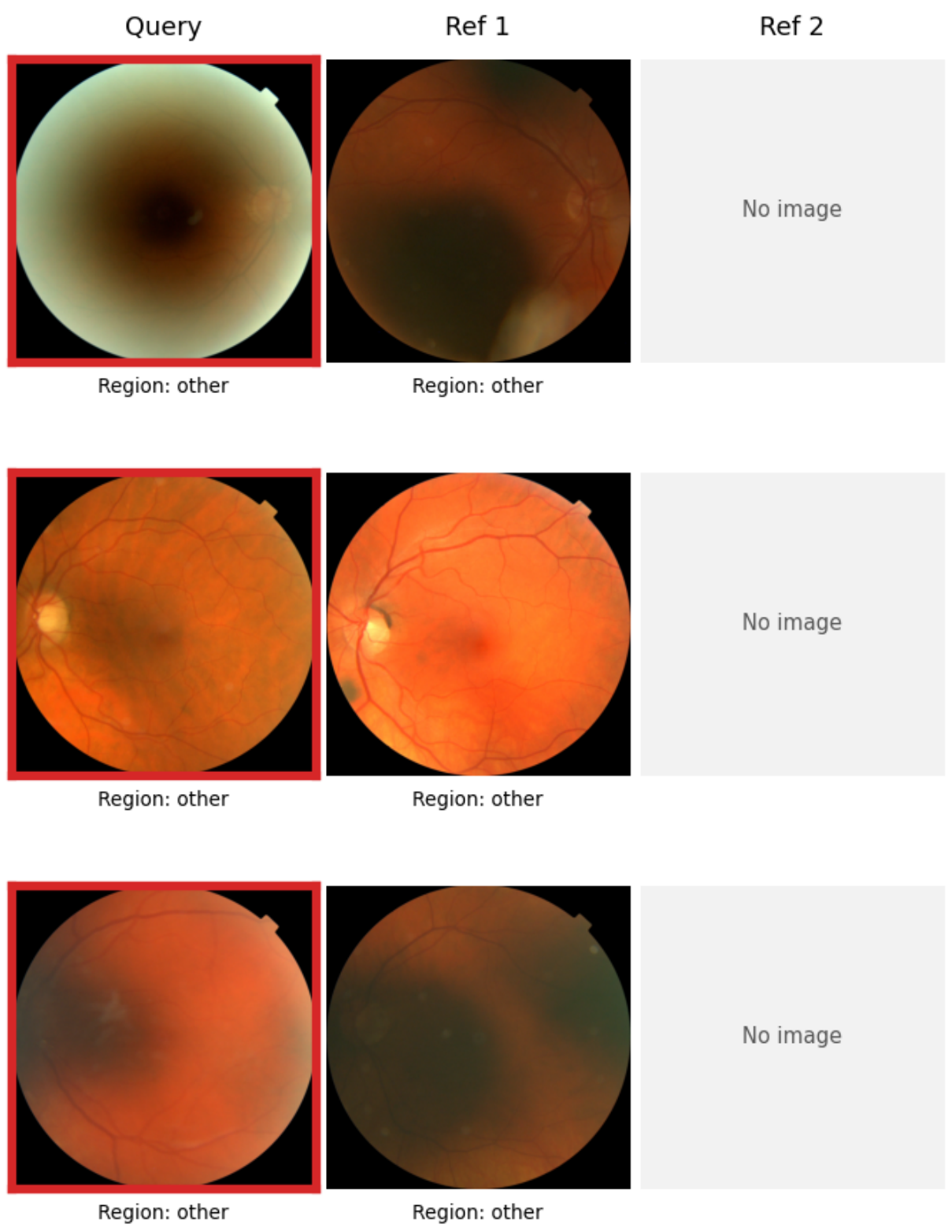

*Supplementary Figure 11. Sample of patient verification false negatives produced by our system in the evaluation of the Retrospective-only scenario on the UK Biobank database. Each group of images presents a simulated incorrect identity assignment consisting in a reference images (Ref 1) and a query image (marked in red) that was deliberately sampled from a different eye (see Evaluation of patient verification and identity retrieval performance). The model failed to flag an incorrect assignment. Images reproduced by kind permission of UK Biobank©.*

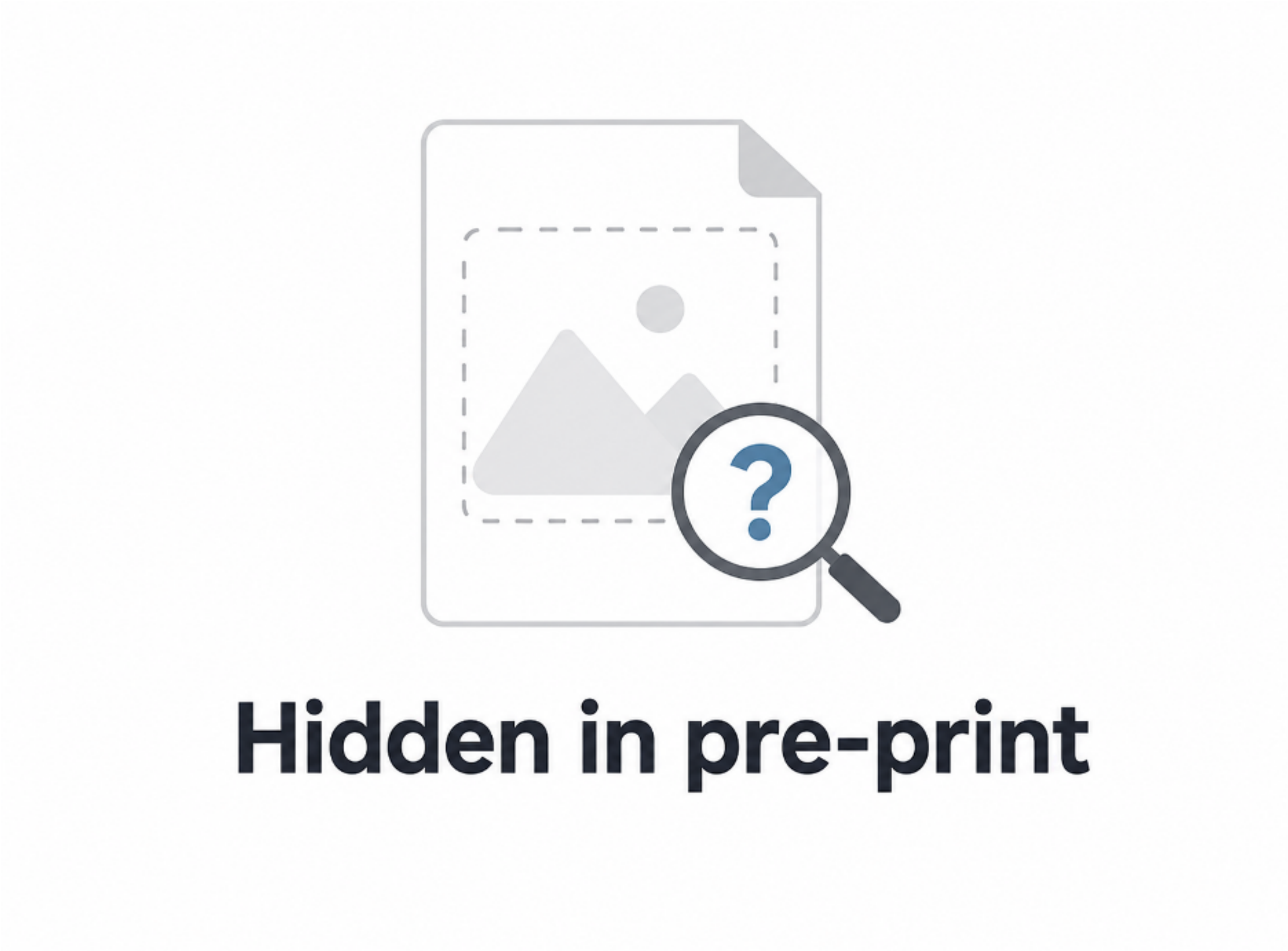

*Supplementary Figure 12. Sample of patient verification false positives produced by our system in the evaluation of the Retrospective-only scenario on the AREDS database. Each group of images (Query + Ref 1 + Ref 2) represents the query image and up to two reference images as presented to the verification system. The three images are assigned to the same patient-eye identity in the reference database, but the verification system flagged an incorrect identity. Reference sets larger than two images were randomly sampled due to space constraints. The vascular patterns partially visible on the images can be used to visually verify that the images correspond to the same eye in most cases. In others, the vascular patterns may be difficult to match.*

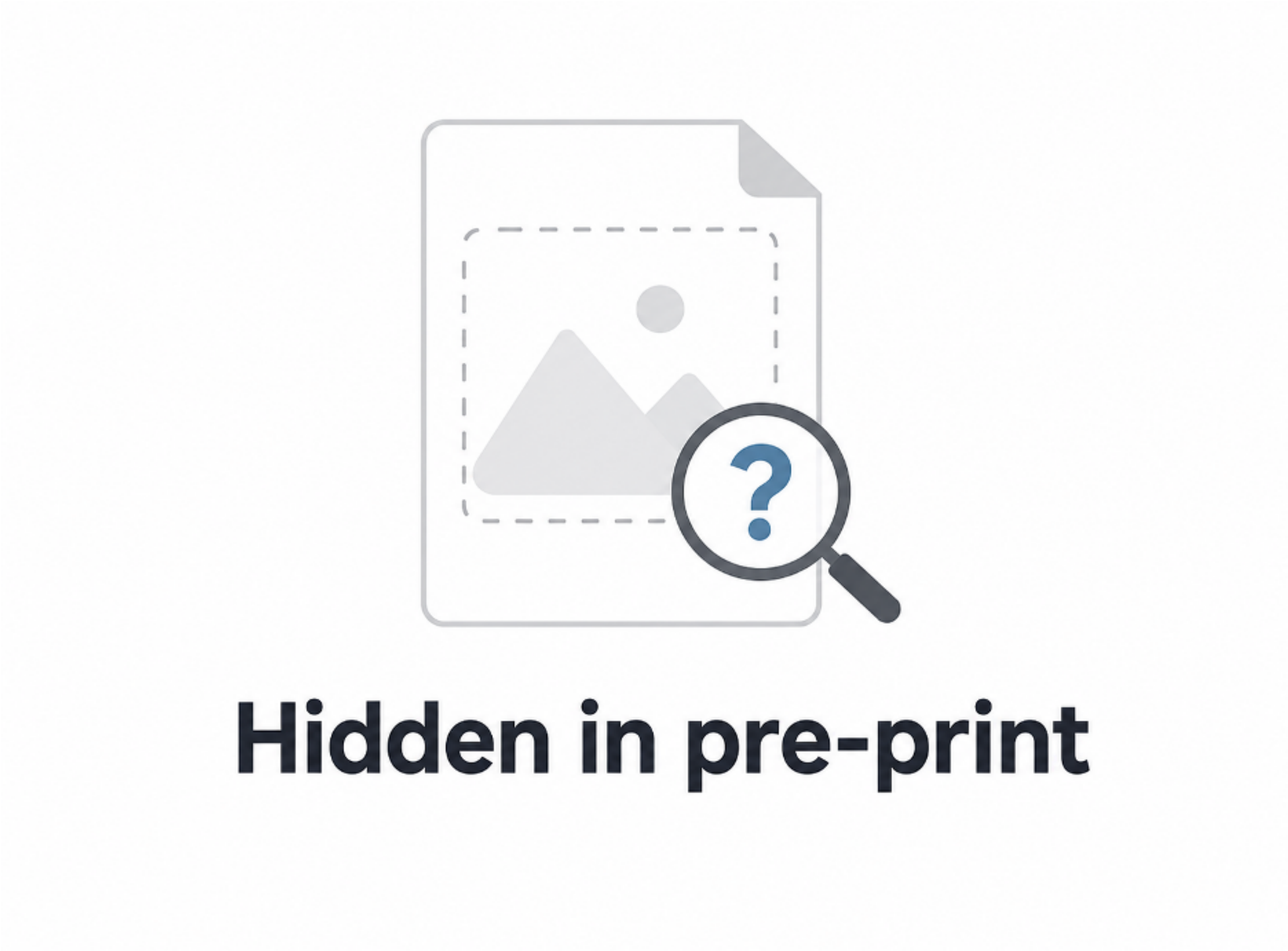

*Supplementary Figure 13. Sample of patient verification false negatives produced by our system in the evaluation of the Retrospective-only scenario on the AREDS database. Each group of three images presents a simulated incorrect identity assignment consisting in two reference images (Ref 1 and Ref 2) and a query image (marked in red) that was deliberately sampled from a different eye (see Evaluation of patient verification and identity retrieval performance). The model failed to flag an incorrect assignment.*

# Patient retrieval failure cases

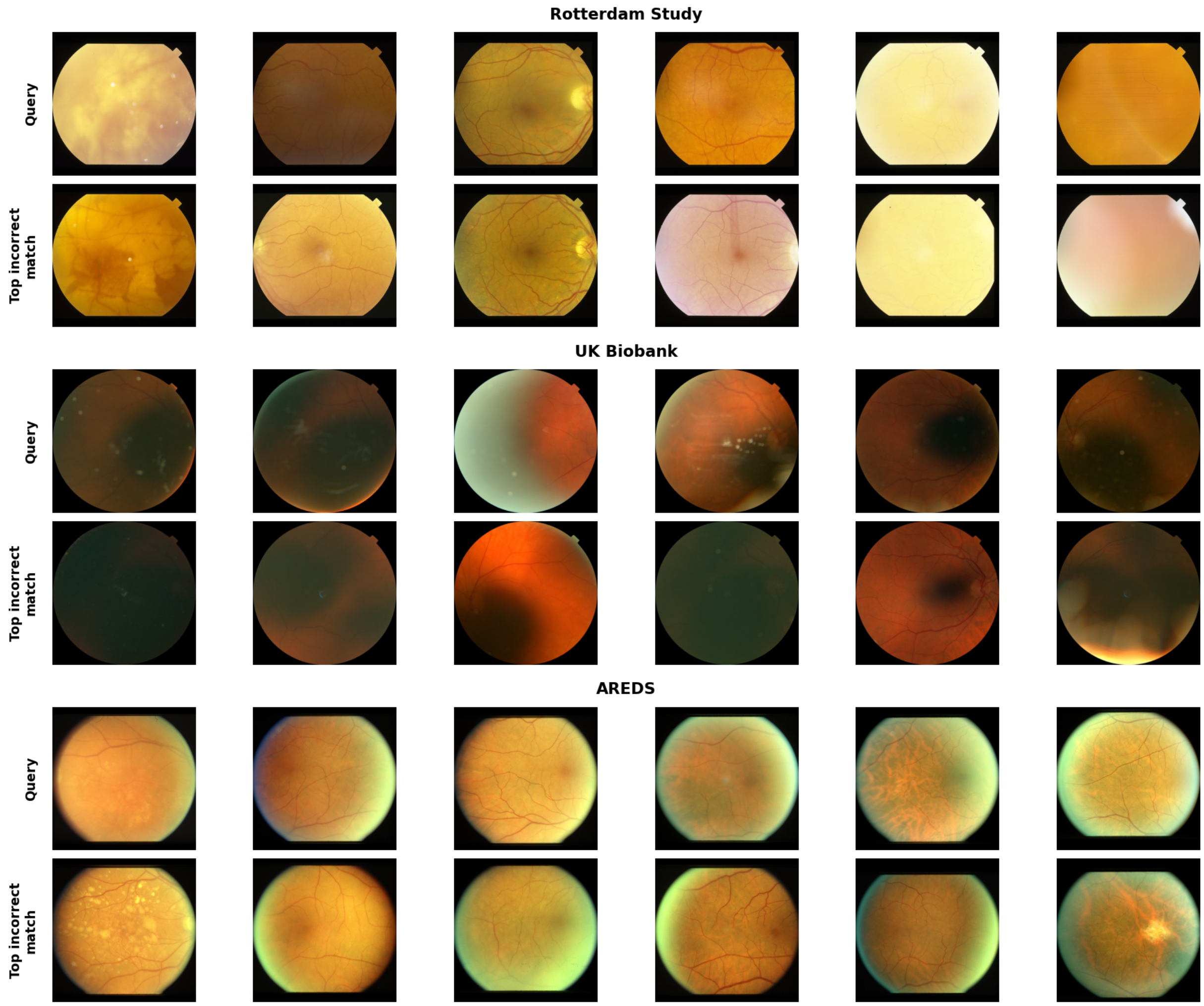

*Supplementary Figure 14. Cases for which the model retrieved an incorrect patient identity during the Retrieval evaluation, for the Retrospective-only scenario on the RS, UKBB and AREDS databases. The Query image identity does not correspond with the model's top match shown below it. Images reproduced by kind permission of UK Biobank©.*

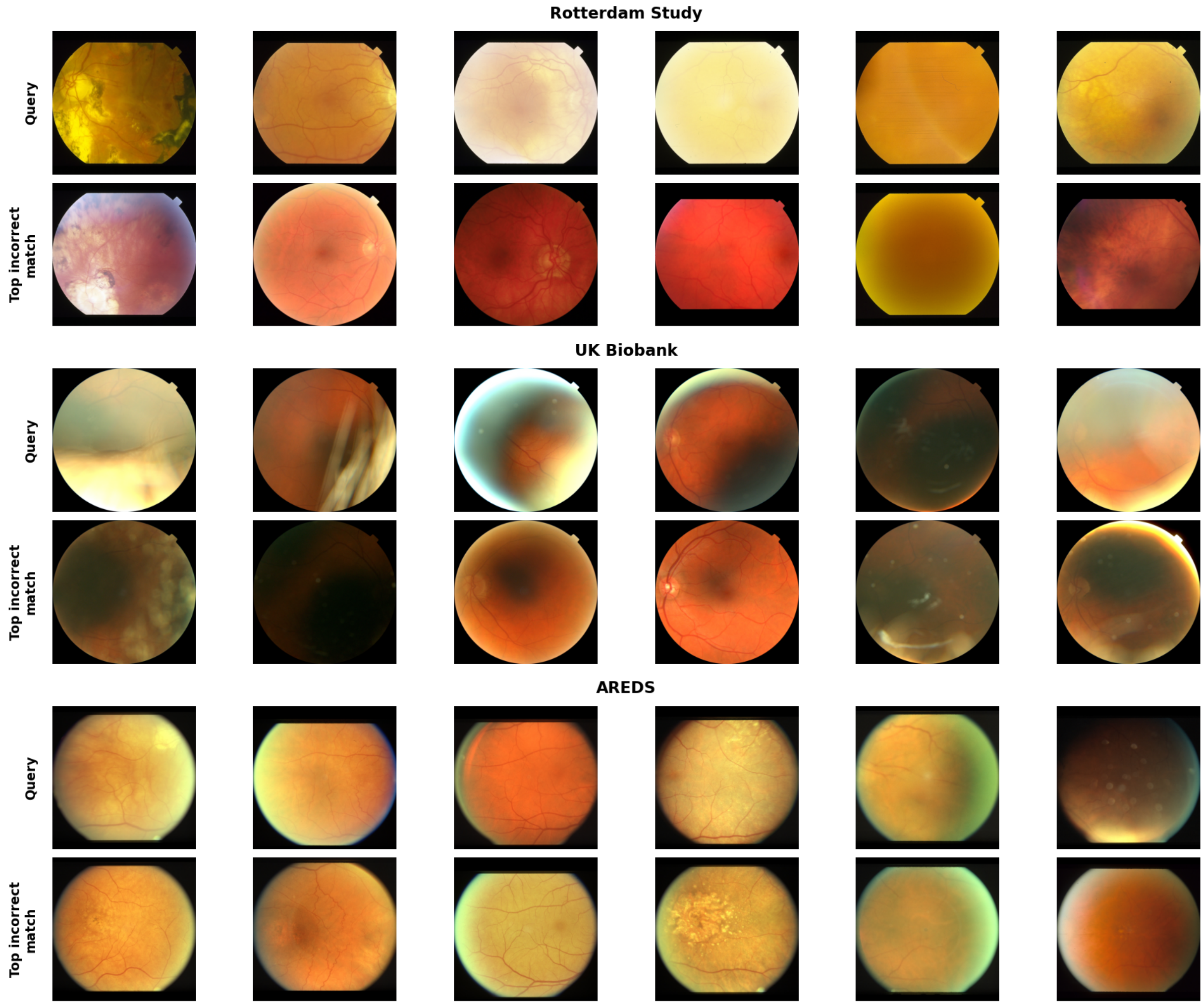

*Supplementary Figure 15. Cases for which the model retrieved an incorrect patient identity during the Retrieval evaluation, for the Retrospective + Prospective scenario on the RS, UKBB and AREDS databases. The Query image identity does not correspond with the model's top match shown below it. Images reproduced by kind permission of UK Biobank©.*

# Appendix F. Members of the VascX Research Consortium

The VascX Research Consortium (in alphabetical order):
Ciara Bergin[3]; Sven Bergmann[1,2,4]; Michael Beyeler[1,2,5]; Dennis Bontempi[1,2]; Sacha Bors[1,2]; Leah Böttger[1,2]; Bogdan Draganski[5,6,7]; Adham Elwakil[3,8]; Györgyi V. Hamvas[9,10]; Janna Hastings[11,12]; Ilaria Iuliani[1,2]; Caroline C.W. Klaver[13,14,15,16]; Ihor Kuras[6,7]; Bart Liefers[13,14]; Ilenia Meloni[3,8]; Sofia Ortin Vela[1,2]; David Presby[1,2]; Ian Quintas[1,2]; José Vargas Quiros[13,14]; Marc Schindewolf[9,10]; Reinier O. Schlingemann[3,17]; Mattia Tomasoni[3,8]; Olga Trofimova[1,2].

1 Department of Computational Biology, University of Lausanne, Lausanne, Switzerland
2 Swiss Institute of Bioinformatics, Lausanne, Switzerland
3 Department of Ophthalmology, University of Lausanne, Fondation Asile des Aveugles, Jules Gonin Eye Hospital, Lausanne, Switzerland.
4 Department of Integrative Biomedical Sciences, University of Cape Town, Cape Town, South Africa
5 Insel University Hospital Bern, Switzerland
6 Department of Neurology, Max Planck Institute for Human Cognitive and Brain Sciences, Germany

7 Department of Clinical Neuroscience, Lausanne University Hospital and University of Lausanne, Switzerland
8 Platform for Research in Ocular Imaging, Fondation Asile des Aveugles, Jules Gonin Eye Hospital, Lausanne, Switzerland.
9 Inselspital, Bern University Hospital, University of Bern, Switzerland.
10 Department for BioMedical Research, Bern University Hospital, University of Bern, Switzerland.
11 Institute for Implementation Science in Health Care, Faculty of Medicine, University of Zurich, Zürich, Switzerland
12 School of Medicine, University of St Gallen, St. Gallen, Switzerland
13 Department of Ophthalmology, Erasmus University Medical Center, Rotterdam, The Netherlands.
14 Department of Epidemiology, Erasmus University Medical Center, Rotterdam, The Netherlands.
15 Department of Ophthalmology, Radboud University Medical Center, Nijmegen, the Netherlands.
16 Institute of Molecular and Clinical Ophthalmology, University of Basel, Switzerland.
17 University Medical Centres, Amsterdam, The Netherlands.